\documentclass[10pt]{article} 
\usepackage[preprint]{tmlr}
\usepackage[utf8]{inputenc} 
\usepackage[T1]{fontenc}    
\usepackage{hyperref}       
\usepackage{url}            
\usepackage{booktabs}       
\usepackage{amsfonts}       
\usepackage{nicefrac}       
\usepackage{microtype}      
\usepackage{xcolor}         
\usepackage{mathtools}
\usepackage{amssymb}
\usepackage{amsmath}
\usepackage{amsfonts}
\usepackage{pifont}
\newcommand{\cmark}{\ding{51}}%
\newcommand{\xmark}{\ding{55}}%

\newcommand{\diff}{\mathrm{d}}
\newcommand{\given}{\,|\,}
\DeclareMathOperator*{\argmin}{arg\,min}
\DeclareMathOperator\erf{erf}

\renewcommand{\mid}{\,|\,}

\usepackage{scalefnt}
\usepackage{enumitem}
\usepackage[skip=0pt]{caption}
\usepackage{subcaption}
\usepackage{tabularx}
\usepackage{array}
\usepackage{multirow}
\usepackage{placeins}
\usepackage{wrapfig}

\usepackage[export]{adjustbox}
\usepackage{gradient-text}

\usepackage{float}

\usepackage{xcolor}
\definecolor{darkblue}{RGB}{0,0,128}
\definecolor{darkyellow}{RGB}{195, 164, 9}
\definecolor{darkgreen}{RGB}{0, 128, 0}
\definecolor{darkred}{RGB}{128, 0, 0}
\definecolor{pink}{RGB}{255, 133, 133}
\definecolor{lightgreen}{RGB}{0, 180, 80}
\definecolor{black}{RGB}{0, 0, 0}

\usepackage{hyperref}
\usepackage{url}  
\hypersetup{
	colorlinks=true,
	linkcolor=darkblue,
	filecolor=darkblue,
	urlcolor=darkblue,
	citecolor=darkblue,
	pdfauthor={},
	pdftitle={}
}

\def\equationautorefname#1#2\null{%
  Eq.#1#2\null%
}
\usepackage{siunitx}
\usepackage{aliascnt}
\usepackage{algorithm}
\usepackage{algpseudocode}

\usepackage{enumitem}
\newcommand{\NA}{---}
\usepackage{tcolorbox}
\tcbset{
  myboxstyle/.style={
    width=\textwidth,colback=blue!4!white,
    left=2pt, right=2pt, top=2pt, bottom=2pt,
  }
}
\tcbset{
  myfancybox/.style={
    enhanced,
    colback=white,
    colframe=blue!70!black,
    boxrule=0.8pt,
    arc=3mm,
    top=1.8cm, 
    overlay={
      \node[fill=blue!70!black,
            text=white,
            rounded corners,
            font=\bfseries,
            anchor=south,
            minimum height=1cm,
            minimum width=6cm] 
            at ([yshift=2mm]frame.north) {\tcbtitle};
    },
  },
}
\newcommand{\eg}{e.g.}
\newcommand{\ie}{i.e.}

\title{Amortized Bayesian Workflow}

\author{%
  \name Chengkun Li \email chengkun.li@helsinki.fi \\ \addr University of Helsinki 
  \AND
  \name Aki Vehtari \email aki.vehtari@aalto.fi \\ \addr ELLIS Institute Finland, Aalto University
  \AND
  \name Paul-Christian Bürkner \email paul.buerkner@tu-dortmund.de \\ \addr TU Dortmund University 
  \AND
  \name Stefan T. Radev \email radevs@rpi.edu \\ \addr Rensselaer Polytechnic Institute
  \AND
  \name Luigi Acerbi \email luigi.acerbi@helsinki.fi \\ \addr University of Helsinki
  \AND
  \name Marvin Schmitt \email mail.marvinschmitt@gmail.com \\ \addr Independent Scientist
}

\def\month{02}  
\def\year{2026} 
\def\openreview{\url{https://openreview.net/forum?id=osV7adJlKD}} 

\begin{document}

\maketitle

\begin{abstract}
Bayesian inference often faces a trade-off between computational speed and sampling accuracy. We propose an adaptive workflow that integrates rapid amortized inference with gold-standard MCMC techniques to achieve a favorable combination of both speed and accuracy when performing inference on many observed datasets. Our approach uses principled diagnostics to guide the choice of inference method for each dataset, moving along the Pareto front from fast amortized sampling via generative neural networks to slower but guaranteed-accurate MCMC when needed. By reusing computations across steps, our workflow synergizes amortized and MCMC-based inference. We demonstrate the effectiveness of this integrated approach on several synthetic and real-world problems with tens of thousands of datasets, showing efficiency gains while maintaining high posterior quality.
\end{abstract}

\section{Introduction}
In many statistical modeling applications, from finance to biology and neuroscience, we often aim to infer unknown parameters $\theta$ from observables $y$ modeled as a joint distribution $p(\theta, y)$ \cite[\eg,][]{raulo_social_2023, seaton_fifty_2023, george_forest_2022,landmeyer_disease-modifying_2020,chen_integrative_2019, malen_atlas_2022, schneider_association_2018,Tsilifis2022}.
The posterior $p(\theta\given y)$ is the statistically optimal solution to this inverse problem, and there are different computational approaches to approximate this target distribution.

Markov chain Monte Carlo (MCMC) methods constitute the most popular family of posterior sampling algorithms and still remain the gold standard for modern Bayesian inference due to their theoretical guarantees and powerful diagnostics \citep{bda3, gelman2020bayesian}.
MCMC methods yield autocorrelated draws conditional on a fixed dataset $y_{\text{obs}}$.
As a consequence, the probabilistic model has to be re-fit for each new dataset, which involves repeating the entire MCMC procedure from scratch.
Modern implementations equip MCMC with state-of-the-art extensions, for example, through Hamiltonian dynamics (HMC; \citealp{nealMCMCUsingHamiltonian2011}), by minimizing the required tuning by users (NUTS; \citealp{hoffmanNoUTurnSamplerAdaptively2014}), or by parallelizing thousands of chains on GPU hardware (ChEES-HMC; \citealp{hoffman2021cheeshmc}). The well-established \emph{Bayesian workflow} \citep{gelman2020bayesian} leverages these tools in an iterative process of model specification, fitting, evaluation, and revision. While powerful, this approach becomes computationally burdensome when applied independently to large collections of datasets.

Differently, \emph{amortized Bayesian inference} (ABI) aims to learn a direct mapping from observables $y$ to the corresponding posterior $p(\theta\given y)$, using flexible function approximators such as deep neural networks \citep{cranmer2020frontier, radev2020bayesflow, greenberg2019automatic,papamakariosNormalizingFlowsProbabilistic2021,wildberger_flow_2023,sharrock_sequential_2024,zammit-mangionNeuralMethodsAmortized2025}.
Amortized inference typically follows a two-stage approach: 
(i) a training stage, where neural networks learn to distill information from the probabilistic model based on simulated examples of observations and parameters $(\theta, y)\sim p(\theta)\,p(y\given\theta)$; and (ii) an inference stage where the neural networks approximate the posterior distribution for an unseen dataset $y_{\text{obs}}$ in near-instant time without repeating the training stage.
In other words: The upfront training cost is \textit{amortized} by negligible inference cost on arbitrary amounts of unseen test data.
Owing to its reliance on simulated data, amortized inference in this form overlaps with \emph{simulation-based inference} \citep{cranmer2020frontier}, which originated from posterior computations for models with intractable likelihood.

\begin{wrapfigure}[14]{r}{0.4\textwidth}
\vspace*{-1.5\baselineskip}
  \begin{center}
    \includegraphics[trim={-0.5cm 14.15cm 26.5cm 0},clip,width=1.0\linewidth]{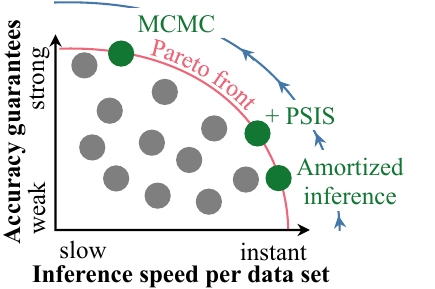}
  \end{center}
  \caption{Our workflow adaptively moves along the Pareto front and reuses previous computations.}
  \label{fig:pareto-front}
\end{wrapfigure}

However, amortized inference lacks the powerful diagnostics and gold-standard guarantees associated with MCMC samplers in the standard Bayesian workflow \citep{gelman2020bayesian}.
Yet, applying a standard workflow is computationally prohibitive at scale. 
In modern Bayesian computation, MCMC and ABI occupy different ends of a Pareto frontier (see \autoref{fig:pareto-front}): the former provides reliable accuracy at high cost, while the latter offers near-instant inference speed with limited per-dataset reliability \citep{hermansCrisisSimulationBasedInference2022,schmitt2023detecting,lueckmann2021benchmarking}.

\begin{figure}[t]
    \centering
    \includegraphics[trim={2.1cm 6.3cm 0 0},clip,width=1.0\linewidth]{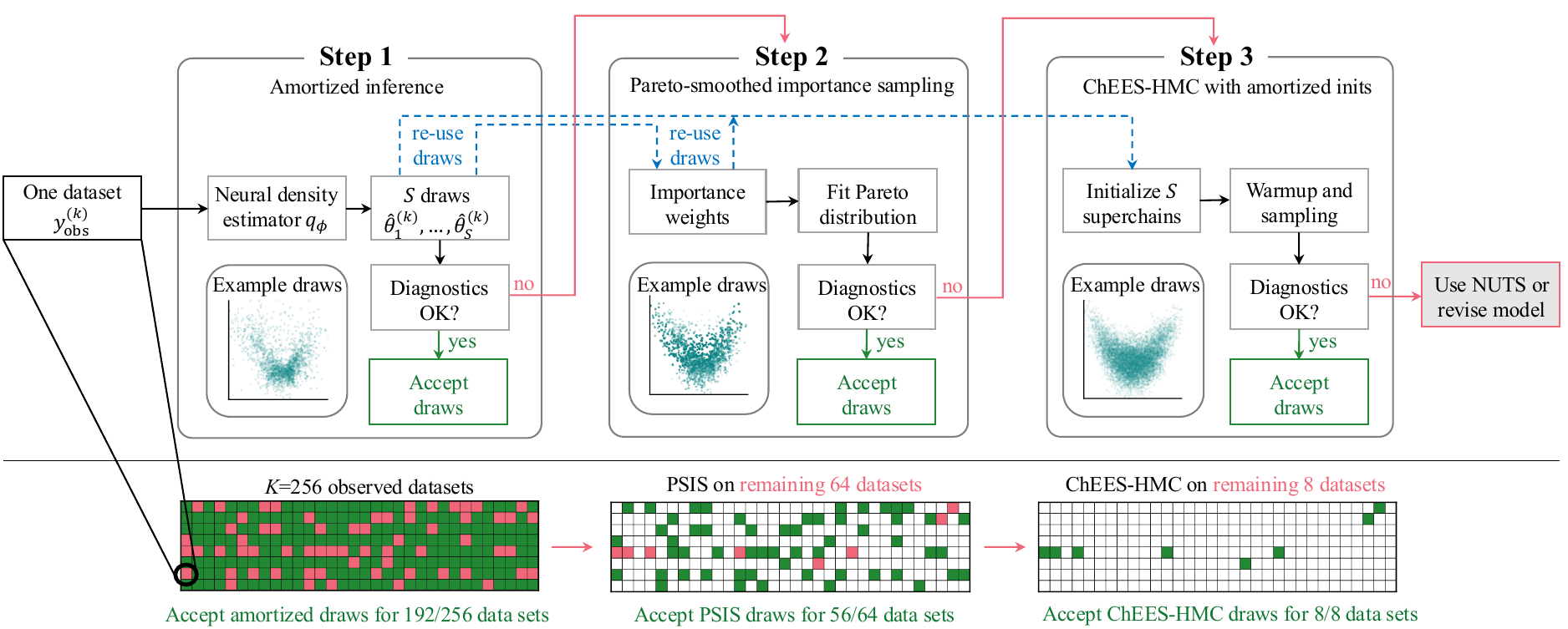}
    \caption{
    Our adaptive workflow leverages near-instant amortized posterior sampling when possible and gradually resorts to slower\textemdash but more accurate\textemdash sampling algorithms.
    As indicated by the blue dashed arrows, we reuse the $S$ draws from the amortized posterior in Step 1 for the subsequent steps in the form of PSIS proposals (Step 2) and initial values in ChEES-HMC (Step 3).
    }
    \label{fig:pull-figure}
\end{figure}

In this paper, we propose an adaptive workflow that consistently yields high-quality posterior draws while remaining computationally efficient. Our proposed workflow \emph{moves along the Pareto front}, enabling fast-and-accurate inference when possible, and slow-but-guaranteed-accurate inference when necessary (see \autoref{fig:pareto-front}). It combines the strengths of ABI and MCMC by incorporating diagnostic checks 
to guide inference decisions and reuse computations wherever possible.
The resulting \emph{amortized Bayesian workflow} therefore offers a principled, scalable, 
and diagnostic-driven approach for efficient posterior inference on many observed datasets; see \autoref{fig:pull-figure} for a conceptual overview.\footnote{The software implementation is available at \url{https://github.com/pipme/amortized-Bayesian-workflow}.}
 To summarize, our contributions are:
\begin{itemize}
    \item Design of—and systematic guidance through—an adaptive Bayesian workflow for accelerating Bayesian inference, which combines the strengths of amortized inference, importance sampling, and MCMC in a theoretically motivated and modular manner.
    \item Empirical validation of the workflow and of its inference speedup, demonstrating the applicability of the workflow on both synthetic and large-scale, real-world problems.
\end{itemize}

\section{Integrating amortized inference into the Bayesian workflow}\label{sec:methods}

Our adaptive workflow starts with neural network training to enable subsequent amortized inference on a large number of unseen datasets\textemdash typically well into tens of thousands.
This training phase is conceptually identical to standalone amortized inference training \cite[\eg,][]{radev2020bayesflow,cranmer2020frontier}.
For the inference phase, however, we develop a principled control flow that guides the analysis.
Based on state-of-the-art diagnostics that are tailored to each step along the workflow, we propose decision criteria to select the appropriate inference algorithm for each observed dataset.
In order to optimize the overall efficiency, our workflow contains mechanisms to reuse previous computations along the way.

\subsection{Training phase: simulation-based optimization}
\label{sec:training_phase}
In ABI, a neural estimator \smash{$q_{\phi}$} with trainable parameters $\phi$ typically minimizes a strictly proper scoring rule $\mathcal{S}$ \citep{gneiting2007strictly, pacchiardiLikelihoodfreeInferenceGenerative2022} in expectation over the joint model $p(\theta, y) = p(\theta)p(y \mid \theta)$,
\begin{equation}\label{eq:min_kl}
    \phi = \argmin_\phi \mathbb{E}_{(\theta, y) \, \sim \, p(\theta, y)}\left[ \mathcal{S} \big(q_\phi(\cdot \mid y),\,\theta \big) \right].
\end{equation}
A popular choice is the logarithmic scoring rule, $\mathcal{S} \big(q_\phi(\cdot \mid y),\,\theta \big) := - \log q_\phi(\theta \mid y)$, which amounts to the forward Kullback-Leibler (KL) objective used for training normalizing flows in ABI \citep{greenberg2019automatic, radev2020bayesflow}.
Score-based formulations that target a time-dependent gradient $\nabla_{\theta_t} \log p(\theta_t \mid y)$ are also possible \citep{sharrock_sequential_2024,gloeckler2024all}.
Since most Bayesian models are generative by design, we can readily simulate $M$ synthetic tuples of parameters and corresponding observations from the joint probabilistic model,
\begin{equation}\label{eq:simulation-based-training}
    (\theta^{(m)}, y^{(m)}) \sim p(\theta, y) \quad\Leftrightarrow\quad \theta^{(m)}\sim p(\theta),\; y^{(m)}\sim p(y\given\theta)\;\;\text{for}\;m=1,\ldots, M,
\end{equation}
which results in the training set $\{(\theta^{(m)}, y^{(m)})\}_{m=1}^{M}$ for optimizing \autoref{eq:min_kl}.
Throughout this paper, we use coupling-based normalizing flows \citep{durkan2019neural, papamakariosNormalizingFlowsProbabilistic2021} as a flexible conditional density estimator $q_{\phi}$ and the forward KL divergence as the training objective. 
However, our proposed workflow is agnostic to the specific choice of generative backbone used for amortization, as long as the model supports efficient sampling (see \autoref{sec:step_1}) and density evaluations (see \autoref{sec:step_2}).

\paragraph{Diagnostics.} 
Since the neural network training algorithm hinges on simulated data, we cannot evaluate the amortized posterior estimator on real data just yet.
However, we can easily simulate a synthetic \emph{test set} $\{(\theta_{\star}^{(j)}, y^{(j)})\}_{j=1}^J$ of size $J$ from the joint model via \autoref{eq:simulation-based-training}.
In this \textit{closed-world} setting, we know which ``true'' parameter vector \smash{$\theta_{\star}^{(j)}$} generated each simulated test dataset $y^{(j)}$. 
A key diagnostic for evaluating the amortized posterior estimator is \emph{simulation-based calibration checking} (SBC; \citealp{talts2018validating,sailynoja2022graphical, modrakSimulationbasedCalibrationChecking2025, yao2023discriminative}). 
Formally, SBC involves (1) defining a test quantity 
\smash{$f: \Theta \times Y \rightarrow \mathbb{R}$} 
(\eg, marginal projections $\theta$ or the log likelihood $p(y\given\theta)$), (2) computing this statistic for the true data-generating parameter \smash{$\theta_{\star}^{(j)}$}, and (3) comparing it to the empirical distribution of the same statistic derived from amortized posterior draws given $y^{(j)}$ \citep{modrakSimulationbasedCalibrationChecking2025}. 
The rank of the true statistic within the posterior draws should be uniformly distributed if the amortized posterior estimator is well-calibrated. 

We recommend assessing uniformity using the graphical approach by \citet{sailynoja2022graphical}, which reveals the type of miscalibration present (e.g., bias or over-/under-dispersion) and is therefore useful for guiding improvements to amortized training. The choice of test quantity in SBC determines the sensitivity of the check; for example, the log-likelihood test quantity is typically more sensitive at detecting discrepancies than marginal projections \citep{modrakSimulationbasedCalibrationChecking2025}; using expressive neural classifiers is also possible \citep{yao2023discriminative}. We further note a trade-off: imposing stricter criteria can improve the fidelity of the amortized estimator but will also tend to reject otherwise practically useful amortized estimators. 

By default, we use marginal projections as the test quantities for SBC and complement SBC checking with \textit{parameter recovery checking}, where parameter estimates are compared against known ground-truth parameters via direct visualization \citep{radev2020bayesflow, radev2023bayesflowjoss}. Parameter recovery checking provides practical insight into whether the learned inverse mapping from $y$ to $\theta$ is effective and helps mitigate a known failure mode of SBC with marginal projections as test quantities, in which the posterior approximation simply recovers the prior. 
We refer to \autoref{app:sec:background} for further details and corresponding pseudocodes.

\paragraph{Note.} Amortized inference lies at the intersection of Bayesian modeling and deep learning, unlocking massive potential for scalable posterior inference.
However, this also comes with the practical challenges inherent to training deep neural networks. 
While a detailed treatment of neural architecture design and optimization exceeds the scope of this paper, practitioners can use established simulation-based inference libraries like \texttt{sbi} \citep{boeltsSbiReloadedToolkit2025} or \texttt{BayesFlow} \citep{radev2023bayesflowjoss}, which provide modern plug-and-play components as well as sensible defaults for a wide range of applications. We summarize a set of best practices and actionable recommendations for training amortized posterior estimators in \autoref{app:sec:best-practices}.

\begin{tcolorbox}[myboxstyle={}]
\textbf{Training phase}: If simulation-based calibration checking and parameter recovery diagnostics pass, proceed to Step~1.
Otherwise, tune the training hyperparameters (\eg, simulation budget, training epochs, learning rate, or neural network architecture) and re-train the amortized network.
\end{tcolorbox}

\subsection{Inference phase: posterior approximation on observed datasets}
Once the amortized estimator is capable of yielding sufficiently accurate posterior draws in closed-world settings (\ie, in-distribution), we use the pre-trained neural network to achieve rapid amortized posterior inference on a total of $K$ observed datasets \smash{$\{y_{\text{obs}}^{(k)}\}_{k=1}^K$}.
Recall that a given pre-trained amortized neural estimator may be perfectly suitable for some real datasets while it is utterly untrustworthy for others.
Therefore, we want to assess on a per-dataset basis whether the amortized posterior draws are trustworthy and should be accepted, or whether we should proceed to a slower algorithm with stronger accuracy guarantees.
The diagnostics in the inference phase are evaluated conditionally on each observed dataset, with the ultimate goal of determining whether the set of current posterior draws is acceptable for that specific dataset.

\subsubsection{Step 1: Amortized posterior draws}
\label{sec:step_1}
\begin{wrapfigure}[14]{r}{0.45\textwidth}
\vspace*{-\baselineskip}
    \centering
    \includegraphics[trim= 3mm 3mm 1mm 3mm, clip, width=\linewidth]{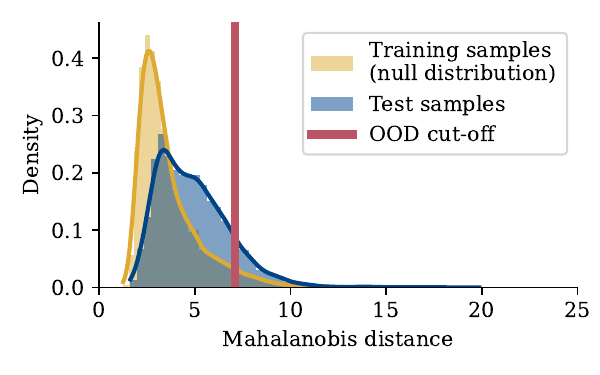}
\vspace*{-15pt}
    \caption{Illustration of our sampling-based hypothesis test that flags OOD datasets (to the right of the OOD cut-off).
    }
    \label{app:fig:ood-test-step-1}
\end{wrapfigure}
We want to exploit the rapid sampling capabilities of the amortized posterior estimator \smash{$q_{\phi}$} as much as possible, as long as the sampled posteriors are trustworthy according to a set of principled diagnostics.
Therefore, the natural first step for each observed dataset \smash{$y_{\text{obs}}^{(k)}$} is to query the amortized posterior and sample $S$ posterior draws $\hat{\theta}_1^{(k)}, \ldots, \hat{\theta}_S^{(k)}\sim q_{\phi}(\theta\given y^{(k)})$ in near-instant time (see \autoref{fig:pull-figure}, first panel).

\paragraph{Diagnostics.}
Like other neural network approaches \citep{yang2024generalized}, amortized inference may yield unfaithful results under distribution shifts \citep{schmitt2023detecting, ward_robust_2022, huang2023learning}.
To address this, we detect whether an observed dataset $y_{\text{obs}}$ is out-of-distribution (OOD) relative to the data-generating process $p(\theta, y)$. We first compute a low-dimensional summary statistic $s(y) \in \mathbb{R}^d$ for each dataset.\footnote{The summary statistic can be either learned by the amortized estimator $q_{\phi}$ in the training phase or be based on domain knowledge.} 
The summary statistics from the training dataset $\{y^{(m)}\}_{m=1}^M$ are used to approximate the Mahalanobis distance by estimating their empirical mean $\mu_s$ and covariance $\Sigma_s$. 
Then, for any test dataset $y$, its Mahalanobis distance to the training set is:
\begin{equation}
    D_M(y) = \sqrt{(s(y) - \mu_s)^\top \Sigma_s^{-1} (s(y) - \mu_s)}.
\end{equation}
We compute $\{D_M(y^{(m)})\}_{m=1}^M$ for all training datasets to establish a frequentist sampling distribution of distances under the null hypothesis (\ie, of in-distribution datasets).
Given a new observed dataset $y_{\text{obs}}$, 
we compare its Mahalanobis distance $D_M(y_{\text{obs}})$ to the empirical distribution of training distances. We define the OOD rejection rule as:
\begin{equation}
    \text{Reject}_{\text{OOD}}(y_{\text{obs}}) = 
    \mathbb{I}\left\{ D_M(y_{\text{obs}}) > \text{Quantile}_{1 - \alpha} 
    \left( \{D_M(y^{(m)})\}_{m=1}^M \right) \right\},
\end{equation}

where $\alpha$ is by default set to 0.05 and we flag datasets whose Mahalanobis distances fall in the right $\alpha$ tail of the empirical training distances as out-of-distribution (see \autoref{app:fig:ood-test-step-1}). The type-I error rate $\alpha$ (false rejection) of this test can be set relatively high to obtain a conservative test that will flag many datasets for detailed investigation in further steps of our workflow.

In a nutshell, this is a sampling-based hypothesis test for distribution shifts, similar in spirit to the kernel-based test proposed by \citet{schmitt2023detecting}.
Since the amortized estimator has no guarantees nor known error bounds for data outside of the empirical support of the joint model $p(\theta, y) $\citep{elsemueller2024sensitivityaware,schmitt2023detecting,frazier2024,elsemüller2025doesunsuperviseddomainadaptation}, we propagate such out-of-distribution datasets to Step 2. It is worth noting that a smaller Mahalanobis distance does not necessarily imply better posterior quality and that this OOD test is only intended to filter out datasets that are most likely to be problematic for the amortized estimator---specifically, those requiring extrapolation outside the ellipsoid defined by the training summary statistics.

\paragraph{Alternative diagnostics.} In addition to the proposed out-of-distribution test, more sophisticated data-conditional diagnostics can further assess the accuracy of amortized posterior draws for individual datasets and enhance the reliability of accepted amortized draws.
Examples include posterior simulation-based calibration checking (posterior SBC; \citealp{sailynoja2025posteriorsbc}) or the local classifier two-sample test (L-C2ST; \citealp{linhart2023lcst}), to name a few. These diagnostics each offer distinct advantages and limitations, but typically require substantially more computation than the OOD test. 

Posterior SBC is conceptually straightforward and offers necessary conditions for the accuracy of amortized posterior samples by assessing consistency. However, it requires additional simulations for each test dataset and requires training the amortized estimator on inputs that effectively double the size of the original observations. L-C2ST, which trains classifiers to distinguish between
$q_{\phi} (\theta \given y)\,p(y)$ and the joint distribution $p(\theta, y)$, provides theoretically sufficient and necessary conditions for amortized inference accuracy. In practice, however, its effectiveness can be very sensitive to several factors, including classifier design choices (e.g., data pre-processing and optimization strategies), classifier calibration, and the relative sizes of the simulation budgets allocated to classifier training and amortized estimator training. 

The choice to apply these additional diagnostics depends on context-specific factors, including the number of observed datasets, the relative computational cost of simulations versus likelihood evaluations,\footnote{For example, if likelihood evaluations are relatively cheap, instead of applying sophisticated diagnostics in Step 1, it is often worthwhile to process directly to Step 2, where Pareto-smoothed importance sampling can offer more informative and powerful diagnostics.} and the dimensionality of the observations. Ultimately, whether amortized posterior draws are deemed acceptable hinges on the accuracy requirements of the specific application. By default, we recommend the OOD test for its simplicity, efficiency, and suitability as a first-line diagnostic.

\vspace{1em}
\begin{tcolorbox}[myboxstyle={}]
\textbf{Step 1}: If the observed dataset passes the OOD test (\ie, Mahalanobis distance is below the threshold), accept the amortized draws; otherwise, proceed to Step~2.
\end{tcolorbox}

\subsubsection{Step 2: Pareto-smoothed importance sampling}
\label{sec:step_2}

In this step, we use Pareto-smoothed importance sampling (PSIS) \citep{vehtari2024psis} to both improve and assess the quality of the amortized posterior draws of datasets which have previously been rejected (see \autoref{fig:pull-figure}, second panel).
Based on the amortized posterior draws from Step 1, PSIS computes importance weights $w_s^{\smash{(k)}} = p(y^{(k)}\given\hat{\theta}_s)\,p(\hat{\theta}_s) / q_{\phi}(\hat{\theta}_s\given y^{(k)})$
for each observed dataset \smash{$y^{(k)}$} (as in default importance sampling).
Then, PSIS fits a generalized Pareto distribution to the largest importance weights, which in turn is used to smooth the tail of the weight distribution \citep{vehtari2024psis}.
Finally, these smoothed importance weights are used for computing posterior expectations and for improving the posterior draws with the sampling importance resampling (SIR) scheme \citep{db1988using}.
While the utility of standard importance sampling for improving neural posterior draws has previously been investigated \citep{dax2023neuralis}, we specifically use the PSIS algorithm, which is self-diagnosing (see \textbf{Diagnostics} below) and therefore better suited for a principled workflow. Further details of PSIS are provided in \autoref{app:sec:background}.

\paragraph{Diagnostics.} 
We use the Pareto-$\hat{k}$ diagnostic to gauge the fidelity of the PSIS-refined posterior draws. Pareto-$\hat{k}$ is the estimated shape parameter of the generalized Pareto distribution and quantifies the tail heaviness of the largest importance weights. 
According to \citet{vehtari2024psis}, for moderate sample size ($S > 2000$), Pareto-\smash{$\hat{k}\leq0.7$} indicates that PSIS estimates are reliable;\footnote{For small sample size ($ S < 2000$), the threshold of Pareto-$\hat{k}$ is $\min (1 -  1 / \log_{10}(S), 0.7)$.} when \smash{$\hat{k}>0.7$}, the minimum sample size for obtaining a reliable Monte Carlo estimate through (Pareto-smoothed) importance sampling rapidly grows infeasibly large in practice, implying that the amortized posterior is a poor proposal for importance sampling correction and the corresponding dataset should be routed to Step~3. This \smash{$\hat{k}$} threshold is consistent with the established practice of using PSIS to improve and assess the quality of posterior approximations obtained from variational inference \citep{yao2018diditwork,dhakaChallengesOpportunitiesHigh2021,zhang2022pathfinder}.

\paragraph{Note.} The posterior estimator in ABI is typically mode-covering since it optimizes the \emph{forward} KL divergence in \autoref{eq:min_kl}. 
When the neural network training is insufficient (\eg, small simulation budget or poorly optimized network), this may lead to overdispersed posteriors.
Fortunately, this tends to err in the right direction, and PSIS can generally mitigate overdispersed \emph{mode-covering} draws in low to moderate dimensions \citep{dhakaChallengesOpportunitiesHigh2021}. 
In contrast, variational inference typically optimizes the \emph{reverse} KL divergence \citep{rezende2015normalizing}, which implies \emph{mode-seeking} behavior that is less favorable for importance sampling.

\begin{tcolorbox}[myboxstyle={}]
\textbf{Step 2}: If Pareto-\smash{$\hat{k}\leq0.7$}, accept the importance sampling results; otherwise, proceed to Step~3.
\end{tcolorbox}

\subsubsection{Step 3: Many-chains MCMC with amortized initializations}
\label{sec:step-3}

If PSIS does not yield satisfactory results, we resort to an MCMC sampling scheme as a safe fallback option.
In our amortized workflow, the MCMC step is augmented by reusing computations from the previous steps as initialization values. 
In principle, this step can incorporate any MCMC algorithm suited to the problem at hand.
Examples include slice sampling for models with non-differentiable likelihoods~\citep{nealSliceSampling2003a}, or HMC~\citep{nealMCMCUsingHamiltonian2011} samplers when gradients are available.

\begin{wrapfigure}[14]{r}{0.40\textwidth}
\vspace*{-0.8\baselineskip}
  \begin{center}
    \includegraphics[trim={0mm 5mm 0mm 5mm},clip,width=0.95\linewidth]{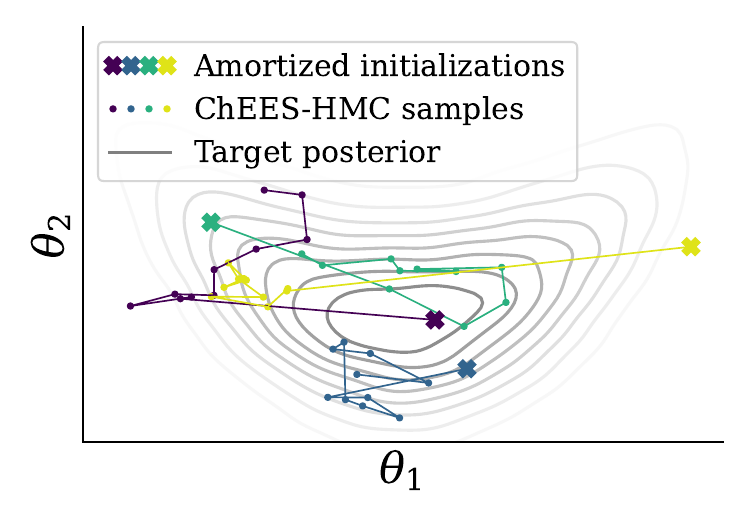}
  \end{center}
  \vspace*{-5pt}
  \caption{We initialize many ChEES-HMC chains with amortized draws.
  }
\label{fig:amortized_init_illustration}
\end{wrapfigure}

In this work, we use the ChEES-HMC algorithm \citep{hoffman2021cheeshmc} as an instantiation of MCMC.
Most notably, ChEES-HMC supports the execution of thousands of parallel chains on a GPU for high-throughput sampling \citep{sountsovRunningMarkovChain2024}.
Amortized posterior draws from previous steps provide a natural and convenient choice for initializing MCMC chains to accelerate convergence (Figure~\ref{fig:amortized_init_illustration}).
%
This approach is conceptually similar to using methods like parallel quasi-Newton variational inference (\ie,\ \emph{Pathfinder}; \citealp{zhang2022pathfinder}) to obtain initial values for MCMC chains.
However, the amortized initial values are drawn in parallel in near-instant time, while Pathfinder requires re-fitting the variational approximation for each new observed dataset.
For the purpose of ChEES-HMC initializations with multimodal posterior distributions, it is again desirable that the amortized posterior draws are typically mass-covering (cf.\ Step 2). See \autoref{app:sec:background} for additional details on the ChEES-HMC algorithm.
\vspace*{-\baselineskip}
\paragraph{Diagnostics.} 
In this last step, we use the nested $\widehat{R}$ diagnostic \citep{margossian2024nestedrhat}, which is specifically designed to assess the convergence of the \textit{many-but-short} MCMC chains.\footnote{In more conventional settings involving long MCMC chains, the standard $\widehat{R}$ diagnostics \citep{gelmanInferenceIterativeSimulation1992,Vehtari2021rhat} can be applied.}
If the diagnostics in this step indicate unreliable inference, we recommend resorting to the overarching Bayesian workflow \citep{gelman2020bayesian} and addressing the computational issues that even persist when using the (ChEES-)HMC algorithm. 
This could involve increasing the number of warmup iterations, using the established NUTS-HMC algorithm \citep{hoffmanNoUTurnSamplerAdaptively2014, carpenter2017stan}, or revising the Bayesian model specification and parametrization.
\begin{tcolorbox}[myboxstyle={}]
\textbf{Step 3}: If (nested) $\widehat{R}$ is below the convergence threshold (\eg, $1.01$), accept the MCMC draws. Otherwise, increase warm-up or revise the model according to the standard Bayesian workflow \citep{gelman2020bayesian}.
\end{tcolorbox}

\subsection{Related work}

Both simulation-based inference and amortized inference have seen rapid progress over the past decade \citep{zammit-mangionNeuralMethodsAmortized2025, cranmer2020frontier,lavin2021simulation}, driven by the need to perform Bayesian inference in complex models with intractable likelihoods \cite[\eg,][]{dingeldein_amortized_2024,wehenkel_simulation-based_2024,zhouEvaluatingSparseGalaxy2024,ghaderi-kangavari_general_2023,vonkrauseMentalSpeedHigh2022,bieringer2021measuring,radev2021outbreakflow}.
These advances have been fueled by modern generative modeling, such as normalizing flows \citep{papamakariosNormalizingFlowsProbabilistic2021,radev2020bayesflow,greenberg2019automatic}, transformers \citep{muller2022transformers,chang2025amortized,whittle2025distribution}, diffusion models \citep{song_score-based_2021,sharrock_sequential_2024,linhart_diffusion_2024,geffner_compositional_2023,gloeckler2024all}, consistency models \citep{song_consistency_2023,schmitt_consistency_2024}, and flow matching \citep{lipman2023flow,wildberger_flow_2023}. Practical software toolkits such as \texttt{BayesFlow} \citep{radev2023bayesflowjoss} and \texttt{sbi} \citep{boeltsSbiReloadedToolkit2025} further make these simulation-based inference techniques accessible to practitioners in user-friendly interfaces.

To address the potential systematic errors of (amortized) neural posteriors, several works propose corrections using importance reweighting schemes \citep{dax2023neuralis,starostinFastReliableProbabilistic2025}, augmented training objectives \citep{delaunoyReliableSimulationBasedInference2022, mishra2025robustsc,Orozco2025,schmitt_leveraging_2024}, or post-hoc corrections \citep{Siahkoohi2023}. 
Simultaneously, hybrid approaches that combine density estimators with MCMC have gained traction \citep{salimansMarkovChainMonte2015,hoffmanNeuTralizingBadGeometry2019, gabrieAdaptiveMonteCarlo2022, midgleyFlowAnnealedImportance2022a, arbelAnnealedFlowTransport2021,cabezasMarkovianFlowMatching2024, greniouxSamplingApproximateTransport2023}. 
These include using variational approximations or learned flows as preconditioners for MCMC 
\citep{hoffmanNeuTralizingBadGeometry2019,cabezasTransportEllipticalSlice2023}, adaptive proposal mechanisms \citep{parnoTransportMapAccelerated2018,gabrieAdaptiveMonteCarlo2022}, 
and initialization strategies to accelerate convergence or improve diagnostics 
\citep{zhang2022pathfinder, wangTargetedAccuracyDiagnostic2023a,starostinFastReliableProbabilistic2025}.

More broadly, automated Bayesian inference has been a central design goal of probabilistic programming systems such as Stan \citep{carpenter2017stan}, PyMC \citep{pymc2023}, (Num)Pyro \citep{binghamPyroDeepUniversal2019,phan2019composable}. These libraries provide general-purpose inference engines---typically gradient-based MCMC and variational inference---that can be applied to a wide range of likelihood-based models and are accompanied by well-developed diagnostic recommendations and workflow guidelines \citep{gelman2020bayesian}. However, they do not natively support amortized inference across many datasets, and inference must be rerun from scratch for each dataset.

Our proposed workflow builds on and complements these lines of work by integrating amortized inference, likelihood-based correction, and many-chain MCMC into a unified, modular, and diagnostic-driven pipeline for accelerating Bayesian inference. It dynamically adapts the inference strategy to the dataset at hand---using amortized posterior draws when they are adequate and escalating to PSIS and MCMC otherwise---thereby improving the robustness of amortized inference and the overall efficiency of posterior computation. This modular design provides a practical foundation for principled amortized inference across diverse data regimes.

\section{Experiments}
In this section, we empirically evaluate the effectiveness of our proposed amortized Bayesian workflow across various synthetic and real-world problems. We also examine how reusing amortized posterior draws in subsequent steps can improve the downstream sampling performance.
The source code to reproduce all experiments is available in the supplementary material.

\subsection{Procedure}

\paragraph{Training settings.} 
For each problem, we begin by training the amortized posterior estimator on simulated parameter-observation pairs (\ie, simulation-based training). We verify that the model performance is satisfactory in a closed-world setting, as diagnosed by simulation-based calibration and parameter recovery checking (see \autoref{sec:training_phase}). Details on diagnostic results, simulation budgets, and training hyperparameters are provided in \autoref{app:sec:experiment}.

\paragraph{Inference settings.}
For the out-of-distribution diagnostics in Step 1, we use the $\alpha = 0.05$ as the rejection threshold. We compute Mahalanobis distances in the summary statistics using 10,000 training simulations. We draw 2,000 posterior samples from the amortized posterior $q_{\phi}$ at Step 1. In Step 2, we correct the amortized draws using PSIS, rejecting draws if Pareto-$\hat{k}>0.7$. Step~3 uses ChEES-HMC with convergence determined by nested $\widehat{R}<1.01$. We run 2048 chains in parallel (16 superchains, each with 128 subchains), with 200 warmup steps and \textit{a single sampling step}, for a total of 2048 posterior draws.

\paragraph{Evaluation metrics.}
To assess the quality of posterior draws from our workflow, we compare them to reference posterior draws using two evaluation metrics: the 1-Wasserstein distance (W1) and the mean marginal total variation distance (MMTV). The W1 distance quantifies the overall discrepancy between full joint distributions. MMTV measures the lack of overlap between marginal distributions and takes value in the range $[0, 1]$; for example, an MMTV value of 0.2 implies that, on average, the approximate posterior draws and reference draws share an 80\% overlap for their marginal distributions. For both metrics, lower values indicate better posterior approximation quality. As a rule of thumb, MMTV values below 0.2 indicate good posterior approximation fidelity \citep{acerbi2020variational, liNormalizingFlowRegression2025}.

\subsection{Applications}
\label{sec:applications}

We apply the proposed workflow to four posterior inference problems, including both simulated benchmarks and real-world experimental datasets. These case studies were chosen to reflect a range of commonly encountered statistical inference scenarios, including classical distributional parameter estimation and analyses of large-scale datasets arising in psychology and cognitive modeling. We describe each problem briefly below, with further details provided in \autoref{app:sec:experiment}.

\paragraph{Generalized extreme value distribution (GEV).}
We consider parameter inference for the generalized extreme value (GEV) distribution, which models the maxima of samples from a distribution family.
Each observation $y_i$ is modeled as:
\begin{equation}
    y_i \sim \text{GEV}(\mu, \sigma, \xi),
\end{equation}
where $\mu \in \mathbb{R}$ is the location, $\sigma \in \mathbb{R}_{>0}$ is the scale, and $\xi \in \mathbb{R}$ is the shape parameter.
We follow the prior specification from \citet{capraniGeneralizedExtremeValue}.
For each dataset, we collect $N = 65$ i.i.d. observations and infer the posterior distribution over the parameter vector $\theta = (\mu, \sigma, \xi)$.
We generate a total of $K=1000$ test datasets by deliberately simulating from a model with a 2$\times$ wider prior distribution to emulate out-of-distribution settings in real applications (see \autoref{app:sec:experiment} for details).

\paragraph{Bernoulli GLM.}  
The Bernoulli generalized linear model (GLM) is a classical model with binary outcomes, included in the SBI benchmark suite \citep{lueckmann2021benchmarking}. Each observation $y_i \in \{0, 1\} $ is modeled as: 
\begin{equation}
    y_i \sim \text{Bernoulli}(\sigma(v_i^\top \theta)),
\end{equation}
where $v_i \in \mathbb{R}^{10}$ is a fixed input vector, $\theta \in \mathbb{R}^{10}$ is the parameter vector, and $\sigma(\cdot)$ denotes the logistic function. 
We generate $K=10,000$ in-distribution test datasets by sampling parameters from the model prior and simulating corresponding observations $\{y_i\}_{i=1}^{100}$ \citep{lueckmann2021benchmarking}.

\paragraph{Psychometric curve fitting.}
Psychometric functions are widely used in perceptual and cognitive science to characterize the relationship between stimulus intensity and the probability of a specific response~\citep{wichmannPsychometricFunctionFitting2001}. We use the overdispersed hierarchical model from \citet{schuttPainfreeAccurateBayesian2016}, 
where the number of correct trials $y_i$ at stimuli level $x_i$ is modeled as:
\begin{equation}
    y_i \sim \text{Binomial}(n_i, p_i), \quad
        p_i \sim \text{Beta}\left( \left( \frac{1}{\eta^2} - 1 \right) \bar{p}_i,\; \left( \frac{1}{\eta^2} - 1 \right)(1 - \bar{p}_i) \right),
\end{equation}
where $n_i$ is the number of trials, $\eta \in [0, 1]$ controls overdispersion, and $\bar{p}_i = \psi(x_i; m, w, \lambda, \gamma)$ is the expected success probability given by the psychometric function $\psi(x; m, w, \lambda, \gamma) = \gamma + (1 - \lambda - \gamma)\, S(x; m, w)$,
where $S$ is a sigmoid function (\eg, cumulative normal), $m$ is the threshold,  $w$ is the width, $\lambda$ is the lapse rate for infinitely high stimulus levels, and $\gamma$ is the guess rate for infinitely low stimulus levels. In total, the model parameters are $\theta = (m, w, \lambda, \gamma, \eta)$. Our empirical evaluation uses 8,526 mouse behavioral datasets from the International Brain Laboratory public database \citep{theinternationalbrainlaboratoryStandardizedReproducibleMeasurement2021}.

\paragraph{Decision model.}
The drift-diffusion model (DDM) is a popular evidence accumulation model for psychological models of human decision making \citep{ratcliff2008}. It describes a two-choice decision task as a stochastic process in which noisy evidence accumulates over time until it reaches one of the decision boundaries. The evolution of the decision variable $z(t)$ is modeled as 
\begin{equation}
    \diff z(t) = v\,\diff t + \sigma\,\diff W(t),
\end{equation}
where $v$ is the drift rate (the average rate of evidence accumulation), 
$\sigma$ is the noise scale, and $W(t)$ denotes a standard Wiener process. 
A decision is made when $z(t)$ reaches either a positive or negative boundary, 
typically placed symmetrically at $\pm a$, where $a$ is the boundary separation. The model also includes a non-decision time parameter $\tau$, 
capturing processes that are not part of the decision process. We adopt the model specification from 
\citet{vonkrauseMentalSpeedHigh2022}, which extends the standard DDM to incorporate experimental condition effects 
via six parameters: $\theta = (v_1, v_2, a_1, a_2, \tau_c, \tau_n$). 
The test datasets consist of 15,000 participants from the online implicit association test (IAT) database \citep{xuPsychologyDataRace2014, vonkrauseMentalSpeedHigh2022}, providing a large-scale, real-world benchmark for Bayesian inference in cognitive modeling.

\subsection{Main Results}

\begin{table}[t]
\caption{
  Summary of our amortized Bayesian workflow across four problems.
  For each step, we report the number of accepted datasets, wall-clock time (minutes), and \textbf{t}ime \textbf{p}er \textbf{a}ccepted dataset (TPA) in seconds. The time for the training phase includes amortized estimator training, simulations, and diagnostics evaluations. 
  The time for Step~1 includes amortized posterior draws and the OOD test. The TPA for Step~1 accounts for both the training phase and Step~1. ``Workflow total'' aggregates the results of our method across all steps.
  As a baseline reference, ``Baseline workflow total'' is an estimate of the total required runtime for ChEES-HMC on all datasets. 
}
\label{tab:exp:results}
\centering
\small
\begin{tabular}{llccc}
\toprule
\textbf{Problem} & \textbf{Step} & \textbf{Accepted datasets} & \textbf{Time (min)} & \textbf{TPA (s)} \\
\midrule
\multirow{4}{*}{\textbf{GEV}} 
& Training phase & \NA & 3 & \multirow{2}{*}{0.4} \\
& Step 1: Amortized inference & $523/1000$ & 0.1 \\
& Step 2: Amortized + PSIS & $357/477$ & 0.8 & 0.1 \\
& Step 3: ChEES-HMC w/ inits & $87/120$ & 11 & 7 \\
& Workflow total (ours) & $967/1000$ & 15 & 0.9 \\
& Baseline workflow total & \NA & 85 & \NA \\
\midrule
\multirow{4}{*}{\textbf{Bernoulli GLM}} 
& Training phase & \NA & 0.8 & \multirow{2}{*}{0.007} \\
& Step 1: Amortized inference & $9519/10000$ & 0.3 \\
& Step 2: Amortized + PSIS & $425/481$ & 0.4 & 0.06 \\
& Step 3: ChEES-HMC w/ inits & $56/56$ & 4 & 4 \\
& Workflow total (ours) & $10000/10000$ & 5 & 0.03 \\
& Baseline workflow total & \NA & 688 & \NA \\
\midrule
\multirow{4}{*}{\textbf{Psychometric curve}} 
& Training phase & \NA & 6 & \multirow{2}{*}{0.06} \\
& Step 1: Amortized inference & $7213/8526$ & 0.4  \\
& Step 2: Amortized + PSIS & $1215/1313$ & 4 & 0.2 \\
& Step 3: ChEES-HMC w/ inits & $69/98$ & 26 & 22 \\
& Workflow total (ours) & $8497/8526$ & 37 & 0.3 \\
& Baseline workflow total & \NA & 2217 & \NA \\
 \midrule
\multirow{4}{*}{\textbf{Decision model}} 
& Training phase & \NA & 85 & \multirow{2}{*}{0.4} \\
& Step 1: Amortized inference & $13498/15000$ & 1 \\
& Step 2: Amortized + PSIS & $827/1502$ & 47 & 3 \\
& Step 3: ChEES-HMC w/ inits & $554/675$ & 526 & 57 \\
& Workflow total (ours) & $14879/15000$ & 659 & 3 \\
& Baseline workflow total & \NA & 11594 & \NA \\
\bottomrule
\end{tabular}
\vspace*{-0.5\baselineskip}
\end{table}

\begin{figure}[t]
    \centering
    \includegraphics[width=\linewidth]{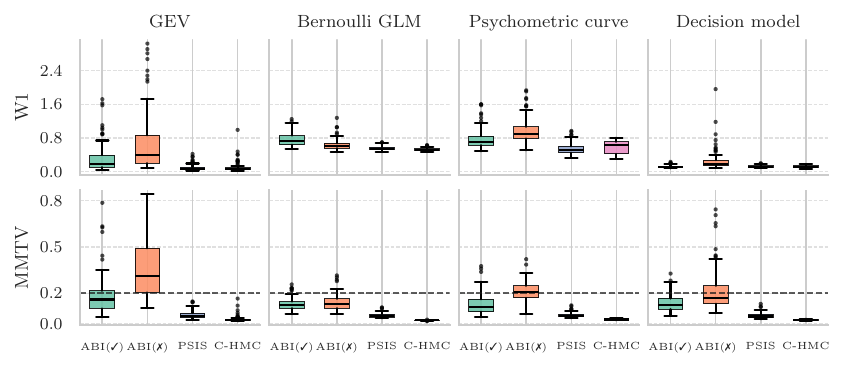}
\caption{Evaluation of posterior draws across four problems based on two metrics: W1 distance (top row) and MMTV distance (bottom row). Lower values indicate better posterior approximation. ABI(\cmark) and ABI(\xmark) denote accepted and rejected draws, respectively, from amortized Bayesian inference in Step 1. PSIS denotes importance-weighted draws accepted in Step 2, and C-HMC denotes draws accepted via ChEES-HMC in Step 3. Metrics are computed on up to 100 datasets for each type of draws.}
    \label{fig:metrics_boxplots}
\vspace*{-\baselineskip}
\end{figure}

\autoref{tab:exp:results} summarizes the performance of the proposed amortized Bayesian workflow across the four problems described in \autoref{sec:applications}. Step~1 (ABI) exhibits extremely low time per accepted dataset (TPA), 
with most of the cost incurred as a one-time expense during the training phase\textemdash including prior simulation, model training, and diagnostic evaluation. Once trained, ABI incurs negligible marginal cost ($\ll 1\text{sec}$) when applied to a new dataset. Datasets flagged as out-of-distribution in Step~1 are forwarded to Step~2 for correction via PSIS. PSIS is highly effective, successfully correcting most rejected amortized draws and substantially reducing the number of datasets requiring full MCMC. Only a small subset of datasets progresses to Step~3, where ChEES-HMC is used for high-fidelity sampling. As the most computationally expensive component, ChEES-HMC is applied selectively, allowing the workflow to retain both accuracy and efficiency. Overall, the amortized workflow completes inference for nearly all datasets.\footnote{A small number of datasets with particularly difficult properties require extended MCMC runs to converge.} 
Compared to using ChEES-HMC for all datasets, our workflow achieves substantial computational savings\textemdash approximately over $5\times$, $120\times$, $60\times$, and $15\times$ faster for the GEV, Bernoulli GLM, psychometric curve, and decision model tasks, respectively.

\autoref{fig:metrics_boxplots} presents the quality of posterior draws using the W1 distance (top row) and MMTV distance (bottom row), comparing draws from each step of the workflow against reference posteriors obtained via well-tuned NUTS. Rejected amortized draws (ABI\xmark) exhibit markedly worse performance than accepted ones (ABI\cmark), confirming the effectiveness of the OOD diagnostics.\footnote{For the Bernoulli GLM, the rejected amortized draws appear of good quality because the test datasets are drawn directly from the same prior used during training (\ie, in-distribution).} PSIS-corrected draws offer accuracy comparable to ChEES-HMC samples, with only a slight decrease in quality. While amortized draws accepted in Step~1 are less accurate than those produced by PSIS or ChEES-HMC, they still provide high-quality approximations across the majority of datasets, as implied by the W1 and MMTV metrics. These results demonstrate that the proposed workflow not only scales efficiently but also consistently produces high-quality posterior estimates.

\subsection{Advantage of amortized initializations for MCMC}
\label{sec:amortized_init_chees}

One major goal of our workflow is to minimize reliance on expensive MCMC by maximizing the reuse of computations. 
Even when ABI and the PSIS refinement fail to yield acceptable posterior draws after Step 2
, we can still leverage the amortized outputs to accelerate MCMC in Step 3.

To evaluate whether amortized posterior estimates remain useful in such cases, we test their effectiveness as initializations for ChEES-HMC chains.
We conduct experiments on 20 randomly selected test datasets that progress to Step~3 of the workflow.
This indicates that both the amortized posterior draws and their Pareto-smoothed refinement are deemed unacceptable, as quantified by Pareto-\smash{$\hat{k}>0.7$} in Step 2.
We compare three initialization methods for ChEES-HMC chains: (1) amortized posterior draws, (2) PSIS-refined amortized draws, and (3) a random initialization scheme similar to Stan \citep{carpenter2017stan}.
We run the chains for varying numbers of warmup iterations, followed by a single sampling iteration. 
As described in \autoref{sec:methods}, we use the nested \smash{$\widehat{R}$} value to gauge whether the chains converged appropriately during the warmup stage, as quantified by the common \smash{$\widehat{R}-1$} threshold of $0.01$ \citep{Vehtari2021rhat}.

\begin{figure*}
    \hfill
    \begin{minipage}{0.60\linewidth}
    \centering    
    \includegraphics[trim={1cm 1.cm 1cm 0cm},width=0.9\columnwidth]{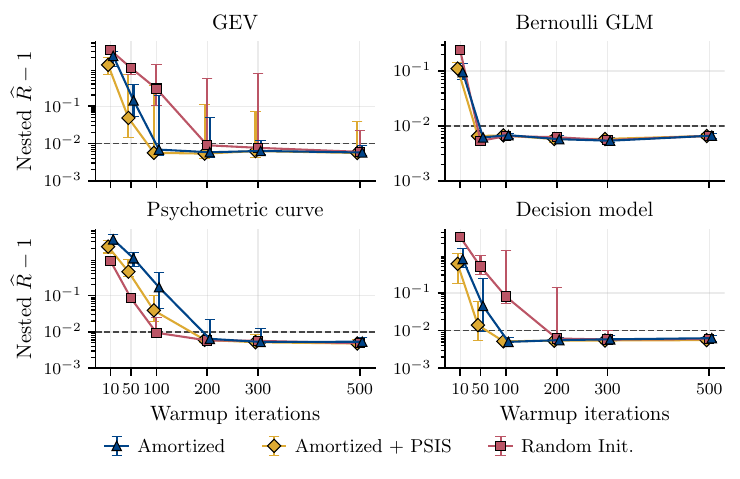}
    \end{minipage}
    \hfill
    \begin{minipage}{0.25\linewidth}
    \caption{
    Using amortized posterior draws as initializations for ChEES-HMC reduces the required warmup in the GEV and decision model tasks. 
    We show median$\pm$IQR across 20 test datasets in Step 3.\label{fig:exp:chees_hmc_inits_comparison}}
    \end{minipage}
\end{figure*}

Figure~\ref{fig:exp:chees_hmc_inits_comparison} shows that amortized posterior draws (and their PSIS-refined counterparts) can significantly reduce the required number of warmup iterations to achieve ChEES-HMC chain convergence, \emph{even though the draws themselves have previously been flagged as unacceptable}. For the GEV problem and the decision model, chains initialized with amortized draws converge faster than those using random initialization. In the Bernoulli GLM, all methods perform similarly. For the psychometric curve model, random initialization leads to faster convergence for the early stage, but amortized draws still reach the convergence threshold at a similar speed at iteration 200, indicating competitive performance. These findings are particularly relevant in the many-short-chains regime, where computational cost is dominated by the warmup phase. For instance, with 2048 parallel chains, every single post-warmup step yields 2048 posterior samples, leading to enormous efficiency gains from shorter warmup.

Overall, these results demonstrate that amortized inference may provide suitable initializations for ChEES-HMC. However, the added benefit of initializing chains with PSIS-refined amortized draws (Step 2) instead of raw amortized draws (Step 1) remains unclear.
While PSIS often accelerates convergence, it occasionally degrades worst-case performance (see upper error bounds for GEV task in \autoref{fig:exp:chees_hmc_inits_comparison}). We further study the impact of initialization for the popular NUTS sampler \citep{hoffmanNoUTurnSamplerAdaptively2014}, with similar results: amortized initializations reduce the required warmup in most cases (see \autoref{app:sec:amortized_init_nuts}).

\section{Discussion}
We presented an adaptive Bayesian workflow to combine the rapid speed of amortized inference with the undisputed sampling quality of MCMC.
Our amortized workflow enables a fundamental shift in the scale and feasibility of Bayesian inference. Applying traditional MCMC (\eg, ChEES-HMC) within a standard Bayesian workflow to every dataset independently 
would require approximately 10 days of GPU computation across our experimental suite. In contrast, our amortized workflow completes inference in half a day, achieving speedups ranging from over $5\times$ to $120\times$ depending on the problem. Crucially, high-quality posterior draws are retained through a cascade of diagnostics and selective escalation to PSIS and MCMC. In conclusion, our workflow efficiently uses resources by (i) applying fast amortized inference when the results are accurate; (ii) refining draws with PSIS when possible; and (iii) amortized initializations of slower but accurate MCMC chains when needed. 


\paragraph{Modularity and practical flexibility.}  
A key strength of the proposed workflow lies in its modular structure, which allows practitioners to tailor each component to the specific constraints and objectives of their application. 
In cases where preliminary analysis or low-latency decision-making is essential (\eg, real-time experimental pipelines) or where likelihood evaluations are computationally expensive, the workflow can operate in a lightweight mode using amortized inference with out-of-distribution rejection alone (\ie, Step 1 in our workflow). 
Conversely, in high-stakes applications where accuracy is paramount, analysts can enforce escalation of all amortized draws through PSIS and, if needed, proceed to full MCMC to guarantee statistical robustness. The choice of MCMC sampler in Step~3 is also fully interchangeable: 
alternative algorithms such as slice sampling, ensemble samplers (\eg, \texttt{emcee}; \citealp{emcee}), 
or NUTS can be substituted if the model is non-differentiable, multimodal, 
or requires richer exploration.

Furthermore, while our paper focuses on the trade-off between wall-clock inference speed and posterior quality, practical deployments may also involve additional factors, such as inference cost (e.g., monetary expense for GPU/CPU hours) and hardware availability. Consequently, the most suitable workflow variant can differ across settings. For example, when GPU resources are limited, 
launching parallel MCMC chains on CPUs offers a practical alternative, making the workflow more accessible for a broader range of users.

\paragraph{Applicability, limitations and future directions.}  
Our proposed workflow targets likelihood-based Bayesian models for which prior predictive simulation and likelihood evaluation are possible. It is most beneficial in repeated-inference regimes (many datasets or frequent re-fits), with moderate effective dimensionality, and when a good amortized estimator can be trained once and subsequently reused. Hence, it is not universally suitable and does not yield inference speedup gains for all Bayesian models.
Training amortized models requires upfront investment in optimization and simulation. In our experiments, we found that default neural network hyperparameter settings, such as normalizing flow architectures, summary network configurations, and optimizer settings, generally yield good performance. 

However, in more challenging cases, such as the GEV problem, adjustments may be necessary, guided by training-phase diagnostics. The simulation burden can be exacerbated in high-dimensional ($\gtrsim 10 $ parameters) or weakly identifiable models, where neural estimators may struggle to approximate complex inverse maps. Alternative amortized inference approaches (see, e.g., \citealp{mittalAmortizedInContextBayesian2025}) could be explored in future work to complement simulation-based amortized inference in such scenarios. In settings where likelihood evaluations are particularly expensive, iterative refinements of the amortized estimator on individual datasets \citep{glockler2022snvi} may also be a practical alternative to likelihood-based corrections in Steps~2 and~3.

Moreover, while our diagnostic for the amortized posterior draws in Step 1 is effective and highly efficient in practice, it remains an imperfect proxy for the true posterior approximation error and can occasionally result in the acceptance of poor-quality amortized draws (cf.\ \autoref{fig:metrics_boxplots}). An additional empirical study in \autoref{app:sec:ood_test_study} shows that (1) the Mahalanobis distance and the posterior quality metrics are positively correlated when OOD datasets are present; (2) however, some low-distance datasets still yield poor metrics, highlighting the limitation that the OOD diagnostic cannot fully guarantee accuracy and motivating enforced escalation to PSIS (Step 2) when higher accuracy is a requirement. Future work could explore even more effective discrepancy measures, potentially tailored to the task at hand.

More broadly, the workflow supports a compelling vision of training amortized models once and reusing them across tasks or studies---a strategy well suited to applications ranging from psychology to computational biology, among others. In such settings, our layered diagnostics and selective escalation are crucial for maintaining reliability and efficiency. This positions the workflow as a practical bridge between amortized inference and traditional Bayesian rigor, enabling scalable yet trustworthy inference.

\subsection*{Acknowledgments}

We thank Alexander Fengler for the help with PyMC implementation of the drift-diffusion model.
CL and LA were supported by the Research Council of Finland (grants number 356498 and 358980 to LA).
AV acknowledges the Research Council of Finland Flagship program: Finnish Center for Artificial Intelligence, and Academy of Finland project 340721. STR is supported by the National Science Foundation under Grant No. 2448380. MS and PB acknowledge support of Cyber Valley Project CyVy-RF- 2021-16, the DFG under Germany’s Excellence Strategy – EXC-2075 -- 390740016 (the Stuttgart Cluster of Excellence SimTech). PB additionally acknowledges the support of DFG Collaborative Research Cluster 391
``Spatio-Temporal Statistics for the Transition of Energy and Transport'' -- 520388526. MS acknowledges travel support from the European Union’s Horizon 2020 research and innovation programme under grant agreements No 951847 (ELISE) and No 101070617 (ELSA), and support from the Aalto Science-IT project. The authors wish to thank the Finnish Computing Competence Infrastructure (FCCI) for supporting this project with computational and data storage resources. The authors also acknowledge the research environment provided by ELLIS Institute Finland.

\bibliography{references.bib}
\bibliographystyle{tmlr}

\clearpage
\newpage
\appendix

\section*{Appendix}

This appendix provides additional details and analyses to complement the main text, included in the following sections:
\begin{itemize}
    \item Background, \autoref{app:sec:background}
    \item Best practices for training amortized estimators, \autoref{app:sec:best-practices}
    \item Experiment details, \autoref{app:sec:experiment}
    \item Additional experimental study of the OOD diagnostic in Step 1, \autoref{app:sec:ood_test_study}
    \item Amortized initialization for NUTS, \autoref{app:sec:amortized_init_nuts}
\end{itemize}

\section{Background}
\label{app:sec:background}
This section provides a concise overview of the diagnostics and algorithms used in our workflow, including simulation-based calibration checking, parameter recovery checking, out-of-distribution diagnostic with Mahalanobis distance, Pareto-smoothed importance sampling, the ChEES-HMC algorithm, and the nested $\widehat{R}$ convergence diagnostic. Pseudocodes are also given for reference.

\paragraph{Simulation-based calibration checking.} 
Simulation-based calibration (SBC; \citealp{talts2018validating, modrakSimulationbasedCalibrationChecking2025}) is a principled technique for assessing the calibration of posterior distributions estimated by Bayesian inference procedures, particularly useful in simulation-based amortized inference settings. SBC is based on the idea that if the posterior $p(\theta \mid y)$ is correctly specified, then the rank of the true parameter $\theta_\star$ among posterior draws should follow a uniform distribution. Formally, SBC defines a test statistic $f: \Theta \times Y \to \mathbb{R}$ (e.g., a component of $\theta$, or the log-likelihood $p(y \mid \theta)$). For each simulated dataset $y^{(j)}$ generated from the joint model $p(\theta, y)$, the test statistic is evaluated at the ground-truth parameter $\theta_\star^{(j)}$ and compared to the same statistic evaluated over posterior samples $\{\theta_s^{(j)}\} \sim q_\phi(\theta \mid y^{(j)})$. The rank of $f(\theta_\star^{(j)}, y^{(j)})$ among $\{f(\theta_s^{(j)}, y^{(j)})\}$ is recorded. Repeating this process for all simulated datasets yields a distribution of rank statistics, which should be uniform under well-calibrated inference. Deviations from uniformity signal systematic bias (e.g., over/under-dispersion) in the posterior approximation. We use the graphical approach proposed by \citet{sailynoja2022graphical} to assess the uniformity of the rank statistics in SBC. This method provides visual diagnostics for identifying systematic biases or miscalibrations in the posterior approximation by plotting the empirical cumulative distribution function (ECDF) and confidence bands (95\%). The pseudocode of SBC is provided in \autoref{alg:sbc}. Examples of SBC checking results using this approach are provided in \autoref{app:sec:experiment}.

\begin{algorithm}[ht]
\caption{SBC diagnostic}
\label{alg:sbc}
\begin{algorithmic}[1]
\Require Joint model $p(\theta,y)$; amortized posterior $q_\phi(\theta \mid y)$; scalar test function $f$; number of simulated datasets $J$ (e.g., 200); posterior draws $S$ (e.g., 1000) 
\For{$j = 1,\dots,J$}
\State Sample $\theta_\star^{(j)} \sim p(\theta)$, $y^{(j)} \sim p(y \mid \theta_\star^{(j)})$
\State Draw $\theta_s^{(j)} \sim q_\phi(\theta \mid y^{(j)})$ for $s = 1,\dots,S$
\State Compute $T_\star^{(j)} = f(\theta_\star^{(j)}, y^{(j)})$, $T_s^{(j)} = f(\theta_s^{(j)}, y^{(j)})$
\State Compute rank $r^{(j)} = \sum_{s=1}^S \mathbb{I} [{T_s^{(j)} < T_\star^{(j)}}] + \mathrm{uniform} (0, \sum_{s=1}^S \mathbb{I} [{T_s^{(j)} = T_\star^{(j)}}])$
\EndFor
\State Compare empirical ranks $r^{(j)}$ against the  $\mathrm{uniform}(0, S)$ distribution using the graphical approach of \citet{sailynoja2022graphical} to identify miscalibration patterns. Alternatively, the uniformity test based on the scalar statistic in Eq.~7 of \citet{modrakSimulationbasedCalibrationChecking2025} can also be applied. 
\end{algorithmic}
\end{algorithm}

\paragraph{Parameter recovery checking.}
Parameter recovery is a complementary diagnostic to SBC and provides a direct visualization of posterior approximation in recovering true generative parameters \citep{radev2020bayesflow, radev2023bayesflowjoss}. The idea is to simulate a collection of datasets $\{y^{(j)}\}$ along with their corresponding ground-truth parameters $\{\theta_\star^{(j)}\}$ from the joint model, and assess whether the posterior distributions $q_\phi(\theta \mid y^{(j)})$ effectively recover these known values. In our workflow, we compare the posterior median extracted from each posterior to the corresponding ground-truth values, along with the median absolute deviation to indicate uncertainty. These comparisons are visualized using scatter plots, with correlation coefficients quantifying the strength of recovery. While not a direct measure of posterior calibration or correctness, parameter recovery provides important practical insight into whether the learned inverse mapping from $y$ to $\theta$ is effective. The pseudocode of the parameter recovery diagnostic is provided in \autoref{alg:param-recovery}. Examples of parameter recovery checking results using this approach are provided in \autoref{app:sec:experiment}.

\begin{algorithm}[ht]
\caption{Parameter recovery diagnostic}
\label{alg:param-recovery}
\begin{algorithmic}[1]
\Require Joint model $p(\theta,y)$; amortized posterior $q_\phi(\theta \mid y)$; number of datasets $J$; posterior draws $S$
\For{$j = 1,\dots,J$}
\State Sample $\theta_\star^{(j)} \sim p(\theta)$, $y^{(j)} \sim p(y \mid \theta_\star^{(j)})$
\State Draw $\theta_s^{(j)} \sim q_\phi(\theta \mid y^{(j)})$ for $s = 1,\dots,S$
\For{each parameter component $k$}
\State Compute posterior median $\hat{\theta}_k^{(j)} = \mathrm{median}_s[\theta_{s,k}^{(j)}]$
\State Optionally compute a dispersion measure (e.g., median absolute deviation) for $\theta_{s,k}^{(j)}$
\EndFor
\EndFor
\State For each $k$, plot $\hat{\theta}_k^{(j)}$ vs.\ $\theta_{\star,k}^{(j)}$ and report correlation
\end{algorithmic}
\end{algorithm}

\paragraph{OOD diagnostic with Mahalanobis distance.}  
The out-of-distribution (OOD) diagnostic used in Step~1 tests whether an observed dataset $y_{\text{obs}}$ falls outside the support of the prior predictive distribution (i.e., the training distribution for the amortized estimator). We work with low-dimensional summary statistics $s(y) \in \mathbb{R}^d$, which are either learned (e.g., via a summary network $h_\psi$) or hand-crafted with domain knowledge. In the latter case, the amortized estimator $q_\phi$ must be trained using these same hand-crafted statistics. The pseudocode for computing the Mahalanobis distance and checking whether an observed dataset $y_{\text{obs}}$ is OOD is provided in \autoref{alg:ood}.

\begin{algorithm}[ht]
\caption{OOD diagnostic with Mahalanobis distance (Step~1)}
\label{alg:ood}
\begin{algorithmic}[1]
\Require Training datasets $\{y^{(m)}\}_{m=1}^M$ (e.g., $M = 10000$); summary statistic function $s(\cdot)$; rejection level $\alpha$ (e.g., $\alpha = 0.05$)
\State Compute $s^{(m)} = s(y^{(m)})$ for all $m$
\State Compute empirical mean $\displaystyle \mu_s = \frac{1}{M} \sum_{m=1}^M s^{(m)}$ and covariance
$\displaystyle \Sigma_s = \frac{1}{M} \sum_{m=1}^M (s^{(m)} - \mu_s)(s^{(m)} - \mu_s)^\top$
\State For each $m$, compute Mahalanobis distance
$\displaystyle D_M(y^{(m)}) = \sqrt{(s^{(m)} - \mu_s)^\top \Sigma_s^{-1} (s^{(m)} - \mu_s)}$

\State Let $T_\alpha$ be the empirical $(1-\alpha)$-quantile of $\{D_M(y^{(m)})\}_{m=1}^M$
\Statex
\Procedure{TestOOD}{$y_{\mathrm{obs}}$}
\State Compute $s_{\mathrm{obs}} = s(y_{\mathrm{obs}})$ and $D_M(y_{\mathrm{obs}})$
\State \Return $\mathbb{I}_{D_M(y_{\mathrm{obs}}) > T_\alpha}$ \Comment{1 = OOD, 0 = in-distribution}
\EndProcedure
\end{algorithmic}
\end{algorithm}

\paragraph{Pareto-smoothed importance sampling.}

Pareto-smoothed importance sampling (PSIS; \citealp{vehtari2024psis}) is a robust method for improving the stability and reliability of importance sampling (IS) estimates. Given a target distribution $p(y \given\theta)\,p(\theta)$ and a proposal distribution $q_{\phi}(\theta)$, with samples $\hat{\theta}_s \sim q_{\phi}(\theta) $, PSIS stabilizes the raw importance weights $w_s = p(y \given\hat{\theta}_s)\,p(\hat{\theta}_s) / q_{\phi}(\hat{\theta}_s\given y)$ by modeling the tail behavior of the importance weights. Specifically, the distribution of extreme importance weights can be approximated by a generalized Pareto distribution (GPD):
\begin{equation}
    p(t \mid u, \sigma, k)= \begin{cases}\frac{1}{\sigma}\left(1+k\left(\frac{t-u}{\sigma}\right)\right)^{-\frac{1}{k}-1}, & k \neq 0 \\ \frac{1}{\sigma} \exp \left(\frac{t-u}{\sigma}\right), & k=0,\end{cases}
\end{equation}
where $k$ is the shape parameter, $u$ is the location parameter and $\sigma$ is the scale parameter. The number of finite fractional moments of the importance weight distribution depends on $k$: a generalized Pareto distribution has $1/k$ finite moments when $k > 0$. To stabilize the importance sampling estimate, the extreme importance weights are replaced with well-spaced order statistics drawn from the fitted generalized Pareto distribution, leading to a more stable and efficient IS estimator. The estimated shape parameter $\hat{k}$ serves as a diagnostic for the reliability of the importance sampling estimate. The pseudocode of PSIS is provided in \autoref{alg:psis}.

\begin{algorithm}[ht]
\caption{PSIS weights and Pareto-$\hat{k}$ diagnostic (Step~2)}
\label{alg:psis}
\begin{algorithmic}[1]
\Require Observed data $y_{\mathrm{obs}}$; log-likelihood $\log p(y \mid \theta)$; log-prior $\log p(\theta)$; amortized posterior $q_\phi(\theta \mid y)$; draws $\theta_s \sim q_\phi$, $s=1,\dots,S$
\For{$s = 1,\dots,S$}
\State $\ell_s = \log p(y_{\mathrm{obs}} \mid \theta_s) + \log p(\theta_s) - \log q_\phi(\theta_s \mid y_{\mathrm{obs}})$
\EndFor
\State $\tilde{\ell}_s = \ell_s - \max_s \ell_s$
\State $w_s = \exp(\tilde{\ell}_s)$
\State Sort ${w_s}$ to obtain $w{(1)} \le \dots \le w{(S)}$
\State Choose tail size $M = \lfloor\min (0.2 S, 3 \sqrt{S})\rfloor$ and define tail ${w_{(S-M+1)},\dots,w_{(S)}}$
\State Fit a GPD to the $M$ largest importance weights ${w_{(S-M+1)},\dots,w_{(S)}}$ and obtain shape estimate $\hat{k}$
\State Replace the $M$ largest weights with smoothed values from the fitted GPD to obtain stabilized weights $\tilde{w}_s$
\State Normalize: $\bar{w}_s = \tilde{w}_s / \sum_{r=1}^S \tilde{w}_r$
\State Use ${\theta_s,\bar{w}_s}$ as weighted PSIS-corrected posterior draws; treat them as reliable if $\hat{k} \le \min (1 -  1 / \log_{10}(S), 0.7)$
\end{algorithmic}
\end{algorithm}

Given the PSIS-stabilized weights ${\bar{w}_s}$ from \autoref{alg:psis}, one can either compute weighted Monte Carlo estimates directly \citep{vehtari2024psis} or apply the SIR procedure in \autoref{alg:sir} to obtain approximately unweighted posterior draws for downstream use (e.g., visualization or MCMC initialization).

\begin{algorithm}[ht]
\caption{Sampling importance resampling (SIR) using PSIS weights}
\label{alg:sir}
\begin{algorithmic}[1]
\Require PSIS-corrected weighted sample $\{(\theta_s,\bar{w}_s)\}_{s=1}^S$; desired number of resampled draws $S'$ (e.g., $S' = S$)
\State Define a categorical distribution on indices $s \in \{1,\dots,S\}$ with probabilities $\bar{w}_1,\dots,\bar{w}_S$
\For{$j = 1,\dots,S'$}
  \State Sample index $I_j \sim \mathrm{Categorical}(\bar{w}_1,\dots,\bar{w}_S)$ 
  \Comment{Typically with replacement; weighted sampling without replacement is useful for generating unique initializations for MCMC chains}
  \State Set $\tilde{\theta}_j = \theta_{I_j}$
\EndFor
\State Return unweighted PSIS-corrected posterior draws $\{\tilde{\theta}_j\}_{j=1}^{S'}$
\end{algorithmic}
\end{algorithm}

\paragraph{ChEES-HMC algorithm.}
The ChEES-HMC algorithm \citep{hoffman2021cheeshmc} is a massively parallel and adaptive extension of Hamiltonian Monte Carlo (HMC) designed to leverage single-instruction multiple-data (SIMD) hardware accelerators such as GPUs. This enables rapid generation of posterior draws following an initial warm-up phase. During warm-up, ChEES-HMC adaptively tunes the trajectory length $T$ and step size $\epsilon$ by maximizing the ``Change in the Estimator of the Expected Square'' (ChEES), a heuristic that serves as a proxy for reducing autocorrelation in the second moments of the Markov chain. ChEES-HMC is particularly suitable for our amortized workflow, as we can easily generate a large number of good starting points (amortized draws) to launch many short MCMC chains. For our experiments, we used ChEES-HMC to run 2048 parallel chains, organized into 16 superchains with 128 subchains each. This configuration is essential for computing the nested $\smash{\widehat{R}}$ diagnostic \citep{margossian2024nestedrhat}, which assesses convergence across a large number of short chains. The pseudocode explaining the use of ChEES-HMC in Step~3 is provided in \autoref{alg:chees-hmc}.

\begin{algorithm}[ht]  
\caption{Use of ChEES-HMC in Step~3}  
\label{alg:chees-hmc}  
\begin{algorithmic}[1]  
\Require Log-posterior density $\log p(\theta, y_\text{obs}) = \log p(y_{\mathrm{obs}} \mid \theta)+\log p(\theta)$ and its gradient w.r.t. $\theta$; number of superchains $K$; number of subchains per superchain $M$; warm-up length $N_{\mathrm{warmup}}$ (e.g.,  $N_{\mathrm{warmup}} = 200$); after warm-up sampling length $N$ (e.g., $N = 1$)
\State Collect $K$ unique amortized posterior draws or PSIS-corrected draws for chain initialization; each of these $K$ draws must have a finite log-posterior density value.
\State For each superchain group $k = 1,\dots,K$, initialize $M$ subchains at the same initial state (total $K \times M$ chains)
\State Run ChEES-HMC warm-up for $N_{\mathrm{warmup}}$ iterations to adapt step size $\epsilon$ and trajectory length $L$  
\State Fix $(\epsilon,L)$ and run $N$ iterations to collect draws from $KM$ parallel chains  
\State Compute nested $\widehat{R}$ for each parameter component of $\theta$
\State If nested $\widehat{R}$ is below a chosen threshold (e.g.\ $<1.01$), accept the combined draws as Step~3 posterior draws; otherwise increase warmup length, use alternative MCMC algorithm (e.g., NUTS-HMC) or revise the model
\end{algorithmic}
\end{algorithm}

\paragraph{Nested $\widehat{R}$ diagnostic.} The potential scale reduction factor $\widehat{R}$ \citep{gelmanInferenceIterativeSimulation1992,Vehtari2021rhat} is arguably the most popular diagnostic for assessing the convergence of MCMC chains. The basic idea is that multiple MCMC chains starting from overdispersed initial points should produce similar Monte Carlo estimators if they have converged, i.e., the impact of initialization vanishes as the chains converge to the stationary distribution. Nested $\widehat{R}$ diagnostic \citep{margossian2024nestedrhat} extends the classical $\widehat{R}$ diagnostic for monitoring convergence of many-short-chain MCMC samplers such as the ChEES-HMC algorithm. It requires organizing chains into $K$ superchains, each consisting of $M$ subchains that share the same initial point. Thus, one can assess convergence through comparing the variability between superchains and the variability within superchains, similar to the standard $\widehat{R}$ diagnostic. We provide the pseudocode for computing the nested $\widehat{R}$ diagnostic in \autoref{alg:nested-rhat} for reference.

\begin{algorithm}[ht]
\caption{Nested $\widehat{R}$ diagnostic}
\label{alg:nested-rhat}
\begin{algorithmic}[1]
\Require Posterior draws $\{\theta^{(nmk)}\}$ after warm-up, where $k \in \{1,\dots,K\}$ (superchains), $m \in \{1,\dots,M\}$ (subchains), $n \in \{1,\dots,N\}$ (draws); scalar function of interest $f$
\State Compute scalar values $f^{(nmk)} \leftarrow f(\theta^{(nmk)})$ for all $n, m, k$.
\State Compute subchain means: $\bar{f}^{(\cdot mk)} \leftarrow \frac{1}{N} \sum_{n=1}^N f^{(nmk)}$
\State Compute superchain means $\bar{f}^{(\cdot \cdot k)} \leftarrow \frac{1}{M} \sum_{m=1}^M \bar{f}^{(\cdot mk)}$ and overall mean $\bar{f}^{(\cdot \cdot \cdot)} \leftarrow \frac{1}{K} \sum_{k=1}^K \bar{f}^{(\cdot \cdot k)}$
\For{each superchain $k = 1, \dots, K$}
    \State Compute between-chain variance $\tilde{B}_k$:
    \State \quad If $M > 1$, $\tilde{B}_k \leftarrow \frac{1}{M-1} \sum_{m=1}^M (\bar{f}^{(\cdot mk)} - \bar{f}^{(\cdot \cdot k)})^2$; else $\tilde{B}_k \leftarrow 0$.
    \State Compute within-chain variance $\tilde{W}_k$:
    \State \quad If $N > 1$, $\tilde{W}_k \leftarrow \frac{1}{M} \sum_{m=1}^M \left( \frac{1}{N-1} \sum_{n=1}^N (f^{(nmk)} - \bar{f}^{(\cdot mk)})^2 \right)$; else $\tilde{W}_k \leftarrow 0$.
\EndFor
\State Compute between-superchain variance: $\widehat{B}_{\nu} \leftarrow \frac{1}{K-1} \sum_{k=1}^K (\bar{f}^{(\cdot \cdot k)} - \bar{f}^{(\cdot \cdot \cdot)})^2$
\State Compute within-superchain variance: $\widehat{W}_{\nu} \leftarrow \frac{1}{K} \sum_{k=1}^K (\tilde{B}_k + \tilde{W}_k)$
\State Return $\text{nested } \widehat{R} \leftarrow \sqrt{\frac{\widehat{W}_{\nu} + \widehat{B}_{\nu}}{\widehat{W}_{\nu}}}$
\end{algorithmic}
\end{algorithm}

\section{Best practices for training amortized estimators}
\label{app:sec:best-practices} 
Amortized inference approaches problems of Bayesian modeling with methods from deep learning. 
While the precise training setup naturally depends on the concrete problem at hand, some general principles have proven help across a wide range of amortized inference applications. We summarize these here as initial guidance for applied practitioners.

\paragraph{Rely on established tooling.}
Modern libraries for amortized inference such as \texttt{sbi} \citep{boeltsSbiReloadedToolkit2025} and \texttt{BayesFlow} \citep{radev2023bayesflowjoss} provide well-tested neural density estimators, training loops, and data pipelines. 
In many cases, their default architectures and optimization settings already yield strong performance without manual tuning. 
First and foremost, we strongly recommend starting from these defaults and only introducing additional complexity if the diagnostics indicate deficiencies.

\paragraph{Monitor the training process.}
In many amortized inference settings, data are generated on the fly by a forward simulation program (see \autoref{eq:simulation-based-training}), and training does not rely on a fixed dataset. In this case, classical data splits into training and validation set are less meaningful, since each minibatch effectively constitutes fresh data from the joint model. 
Nonetheless, it is still important to monitor training progress with multiple signals, such as the training loss, calibration diagnostics, or summary statistics of posterior samples. These checks help assess whether the model continues to improve or has already reached a performance plateau after a short period.
In other settings, amortized inference may be trained on a \emph{fixed} set of simulated data, for example due to an expensive simulator or precomputed datasets. 
In such cases, holding out a validation set is strongly recommended to detect overfitting and guide selection of the amortized estimator.

\paragraph{Track multiple performance signals.}
Regardless of whether data are simulated on the fly or fixed, we recommend monitoring multiple indicators during training (e.g., after each epoch). 
Loss curves provide a coarse signal of optimization progress but are often insufficient on their own. 
For example, normalizing flows are usually trained with a negative log-likelihood loss, which does not account for mode coverage in multi-modal posterior distributions.
Complementary diagnostics such as simulation-based calibration, parameter recovery, or posterior predictive checks on held-out datasets offer more direct insight into posterior quality and failure modes.

\paragraph{Assess and adjust model expressiveness.}
When training stagnates or diagnostics indicate systematic errors, the limiting factor is often model expressiveness rather than optimization details. 
Underexpressive models may show symptoms such as poor parameter recovery, persistent miscalibration, or posterior collapse toward the prior (i.e., data insensitivity), even when the training loss decreases. 
A pragmatic strategy is to begin with an overly expressive architecture (i.e., many trainable weights) to establish a performance baseline, and then gradually simplify the model. 
Conversely, if diagnostics remain unsatisfactory, increasing model capacity is often more effective than tuning hyperparameters of the optimizer.

\paragraph{Accept that training is iterative.}
Even with modern tooling, amortized inference training may require many iterations during development, especially for complex or weakly identifiable models. 
The objective is not to find a universally optimal neural architecture, but to reach a regime where the amortized posterior is reliable enough to enter the adaptive workflow proposed in this paper, where subsequent diagnostics and correction steps can take over.

\section{Experiment details}
\label{app:sec:experiment}
In this section, we provide additional experimental details omitted from the main text for brevity.

\paragraph{Evaluation metrics.}
To assess the quality of posterior approximations produced by each step of the workflow, we compare them against reference posterior draws obtained via a well-tuned 
No-U-Turn Sampler (NUTS). Specifically, we precomputed NUTS-based posterior samples for a subset of \num{5000} test datasets, 
which serve as a ground-truth reference for evaluation.\footnote{For the generalized extreme value distribution problem, \num{1000} reference datasets were used, corresponding to the total number of test datasets.} We then evaluate the 1-Wasserstein distance (W1) and the mean marginal total variation (MMTV) distance on up to 100 datasets from each inference step: Step~1 (amortized inference), 
Step~2 (amortized + PSIS), and Step~3 (ChEES-HMC with amortized initializations). These metrics are reported in the main text (Figure~5).

\paragraph{Neural network architecture for amortized inference.} For all experiments, we use a coupling-based normalizing flow implemented in \texttt{BayesFlow} \citep{radev2023bayesflowjoss}. The flow consists of 6 transformation layers, each comprising an invertible normalization, two affine coupling transformations, and a random permutation between elements. Before entering the coupling flow network as conditioning variables, the observed dataset $y$ is encoded into a lower-dimensional summary statistic $h_{\psi}(y)$ via a summary network $h_{\psi}$. This summary network is implemented either as a DeepSet architecture \citep{zaheerDeepSets2017} or a SetTransformer \citep{leeSetTransformerFramework2019}, depending on the problem setting. Both architectures are designed to handle permutation-invariant data structures. For the Bernoulli GLM, we bypass the summary network and directly use the known 10-dimensional sufficient statistics \citep{lueckmann2021benchmarking}. The specific choice of summary network for each application is described in the respective problem descriptions below. 


\paragraph{Training-phase optimization.} For all problems, the neural network is optimized via the AdamW optimizer \citep{loshchilovDecoupledWeightDecay2019} with weight decay of $10^{-3}$ and a cosine decay learning rate schedule (initial learning rate of $2.5 \times 10^{-4}$, a warmup target of $5 \times 10^{-4}$, $\alpha=0$) as implemented in Keras \citep{chollet2015keras}. A global gradient clip norm of $1.5$ is applied. Training is performed with a batch size of $512$ for 300 epochs,\footnote{For Bernoulli GLM, we only train for 100 epochs.} with cosine decay steps set to the product of batch size and epochs. 
A held-out validation set is used to monitor optimization and select the best-performing model checkpoint.

\paragraph{Space transformation.} 
Following standard practice in Bayesian computation (e.g., PyMC; \citealp{pymc2023}, Stan; \citealp{carpenter2017stan}), we transform parameters to an \emph{unconstrained} space for inference. The amortized neural estimator is trained to estimate parameters in this unconstrained space. PSIS operates independently of the parameterization and thus remains unaffected by this transformation. ChEES-HMC also performs inference in the unconstrained space. All evaluation metrics (W1 and MMTV distances) are computed in this space. However, parameter recovery and simulation-based calibration plots are shown in the original constrained space for better interpretability.

\paragraph{Computing infrastructure and software.}
For all applications, the full workflow---including amortized training, inference, 
diagnostics, Pareto-smoothed importance sampling, and ChEES-HMC sampling---was conducted on a single NVIDIA V100 GPU (32GB), 8 cores of an AMD EPYC 7452 processor, and 8-16GB RAM. For runtime details across experiments, refer to Table 1 in the main text. The core code base was built using \texttt{BayesFlow} \citep{radev2023bayesflowjoss} (MIT license), PyMC \citep{pymc2023} (Apache-2.0 license), ArviZ \citep{arviz_2019} (Apache-2.0 license) and JAX \citep{jax2018github} (Apache-2.0 license). We used the implementation of ChEES-HMC provided by TensorFlow Probability \citep{tensorflow_distributions} (Apache-2.0 license).

Below, we provide details for each problem to complement the main text.

\subsection{Generalized extreme value distribution}

\paragraph{Problem description.}

Following \citet{capraniGeneralizedExtremeValue}, the prior distribution for the parameters of the generalized extreme value distribution (GEV) is defined as: 
\begin{equation}
\begin{aligned}
    \mu & \sim \mathcal{N}(3.8, 0.04)\\
    \sigma & \sim \text{Half-Normal}(0, 0.09) \\
    \xi & \sim \text{Truncated-Normal}(0, 0.04)\;\text{with bounds}\;[-0.6, 0.6].
\end{aligned}
\end{equation}

\paragraph{Simulation budgets.} 
We use \num{10000} simulated parameter–observation pairs for training the amortized estimator, \num{1000} for validation, and \num{200} for training-phase diagnostics, including parameter recovery and simulation-based calibration.

\paragraph{Summary network.}
We use a DeepSet as the summary network. The dimensionality of the learned summary statistics is 16. The DeepSet has a depth of 1, uses a \emph{mish} activation, max inner pooling layers, 64 units in the equivariant and invariant modules, and 5\% dropout.

\begin{figure}[t]
    \centering
    \begin{subfigure}[t]{0.70\linewidth}
        \includegraphics[width=1.0\linewidth]{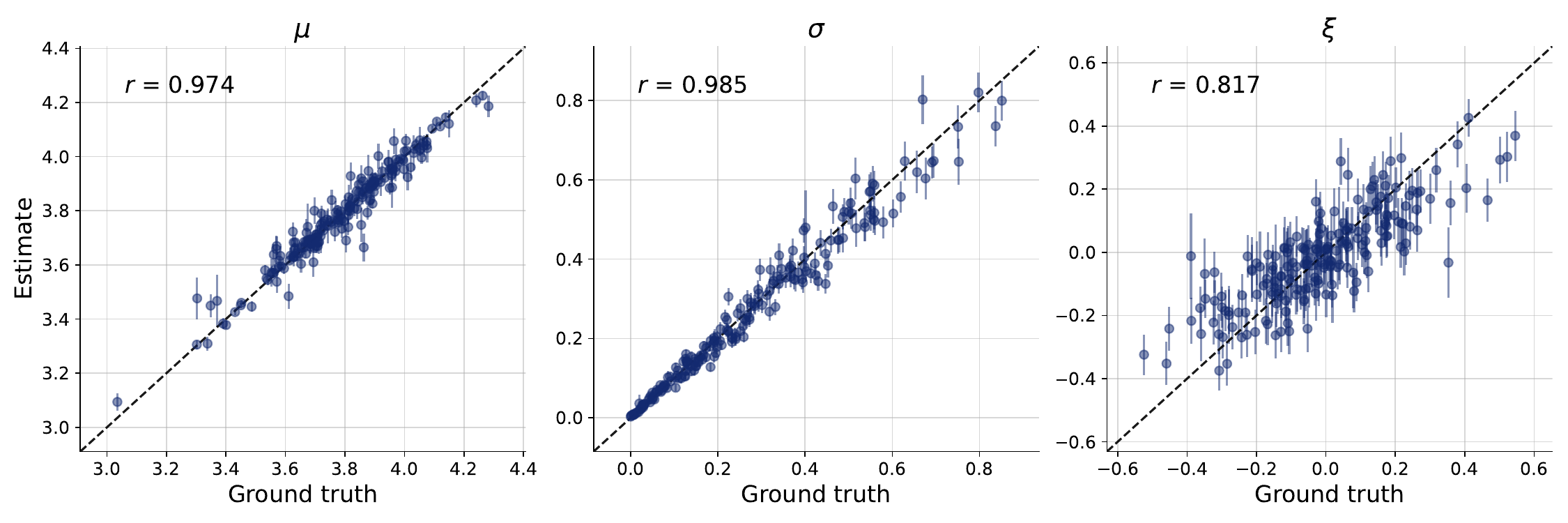}
        \caption{Parameter recovery checking.}
        \label{app:fig:recovery-GEV}
    \end{subfigure}
    \\
    \begin{subfigure}[t]{0.70\linewidth}
        \centering\includegraphics[width=1.0\linewidth]{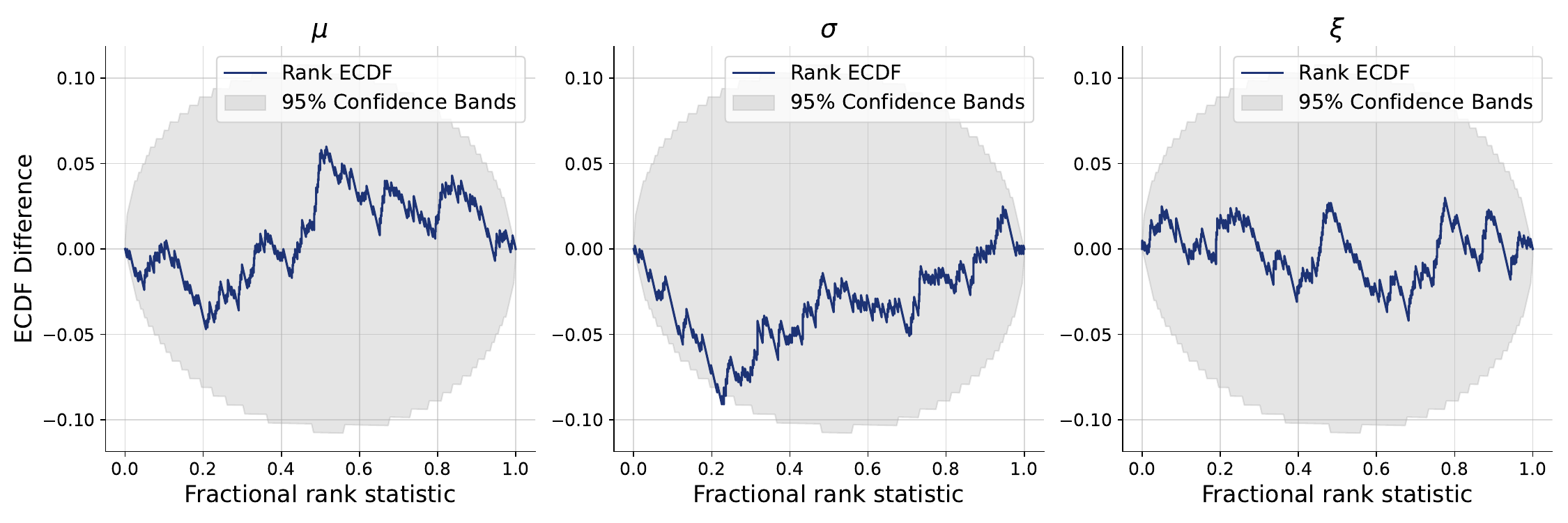}
        \caption{Simulation-based calibration checking.}
    \label{app:fig:calibration_ecdf-GEV}
    \end{subfigure}
    \caption{Training-phase diagnostics for the GEV problem. The parameter recovery is strong for the parameters $\mu, \sigma$, and good for the shape parameter $\xi$. Simulation-based calibration checking indicates good calibration for all parameters. Parameter recovery and simulation-based calibration checking indicate acceptable convergence of the amortized posterior estimator.}
    \label{app:fig:training_diagnostics-GEV}
\end{figure}

\paragraph{Training-phase diagnostics.} The closed-world diagnostics (parameter recovery and simulation-based calibration checking) in \autoref{app:fig:training_diagnostics-GEV} indicate that the neural network training has successfully converged to an acceptable posterior estimator within the scope of the training set.

\paragraph{Test datasets.}
In order to emulate distribution shifts that arise in real-world applications while preserving the controlled experimental environment, we simulate the ``observed'' datasets from a joint model whose prior is $2\times$ wider (i.e., with $4\times$ the variance) than the model used during training.
More specifically, the prior is specified as:
\begin{equation}
\begin{aligned}
    \mu & \sim \mathcal{N}(3.8, 0.16)\\
    \sigma & \sim \text{Half-Normal}(0, 0.36) \\
    \xi & \sim \text{Truncated-Normal}(0, 0.16)\;\text{with bounds}\;[-1.2, 1.2].
\end{aligned}
\end{equation}

\subsection{Bernoulli GLM}

\paragraph{Problem description.} Following \citet{lueckmann2021benchmarking}, we set the prior for $\theta$ as:
\begin{equation}
\begin{aligned}
    \theta \sim \mathcal{N} \left(0, 
    \begin{bmatrix}
    2 & 0 \\
    0 & (F^\top F)^{-1}
    \end{bmatrix}
    \right),
\end{aligned}
\end{equation}
where the matrix $F$ is defined such that 
$F_{i, i-2}=1$, 
$F_{i, i-1}=-2$, 
$F_{i, i}=1+\sqrt{\frac{i-1}{9}}$, 
and $F_{i, j}=0$ otherwise, for $1 \leq i, j \leq 9$. The task duration is set to $T=100$, with fixed input vectors $\{v_i\}_{i=1}^{100}$, where each $v_i \in \mathbb{R}^{10}$. Corresponding observations are denoted by $\{y_i\}_{i=1}^{100}$. Further details can be found in \citet{lueckmann2021benchmarking, goncalvesTrainingDeepNeural2020a}.

\paragraph{Simulation budgets.} 
We use \num{10000} simulated parameter–observation pairs for training the amortized estimator, \num{1000} for validation, and \num{200} for training-phase diagnostics, including parameter recovery and simulation-based calibration.

\paragraph{Summary network.} For Bernoulli GLM, the 10-dimensional sufficient summary statistic for each dataset can be computed as $Vy^\top$ where $y = [y_1, \cdots, y_{100}]$ and $V = [v_1, \cdots, v_{100}]$. We therefore use this summary statistic for amortized training directly without relying on a separate summary neural network.

\begin{figure}[t]
    \centering
    \begin{subfigure}[t]{\linewidth}
        \includegraphics[width=1.0\linewidth]{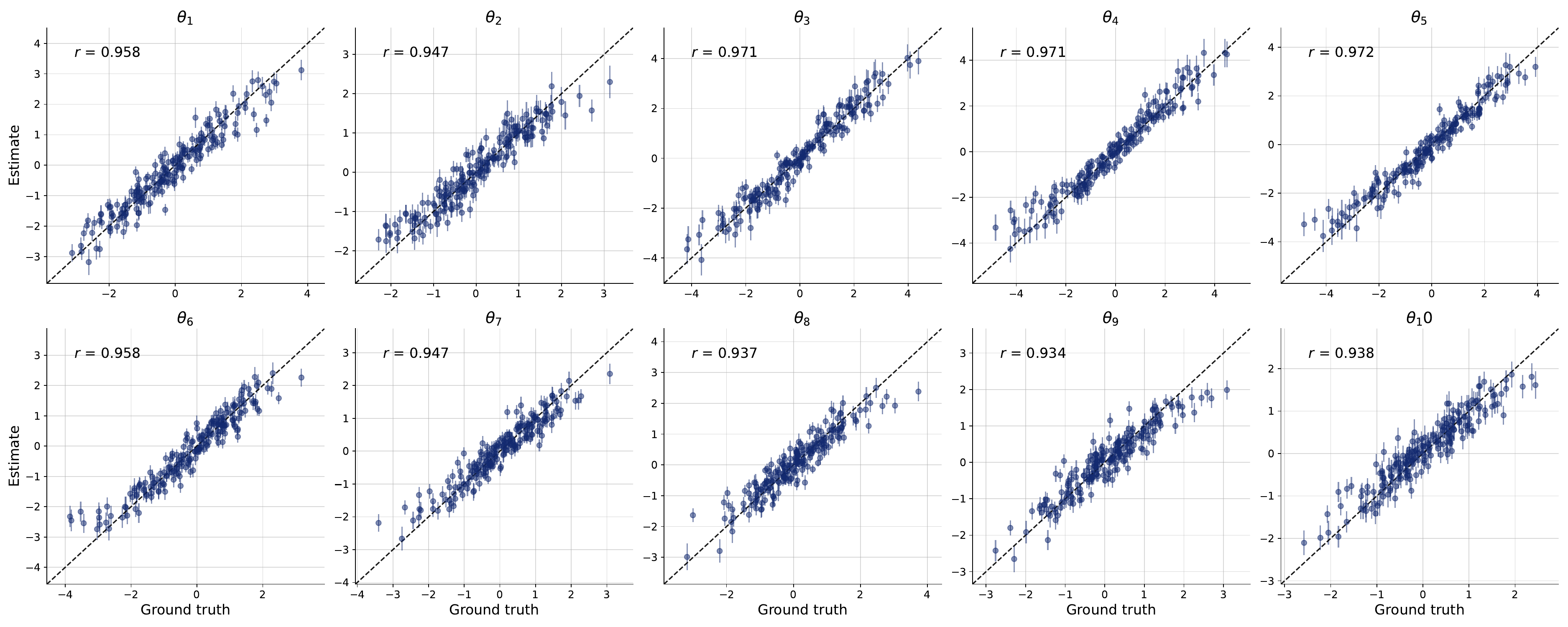}
        \caption{Parameter recovery checking.}
    \end{subfigure}
    \\
    \begin{subfigure}[t]{\linewidth}
        \centering
        \includegraphics[width=1.0\linewidth]{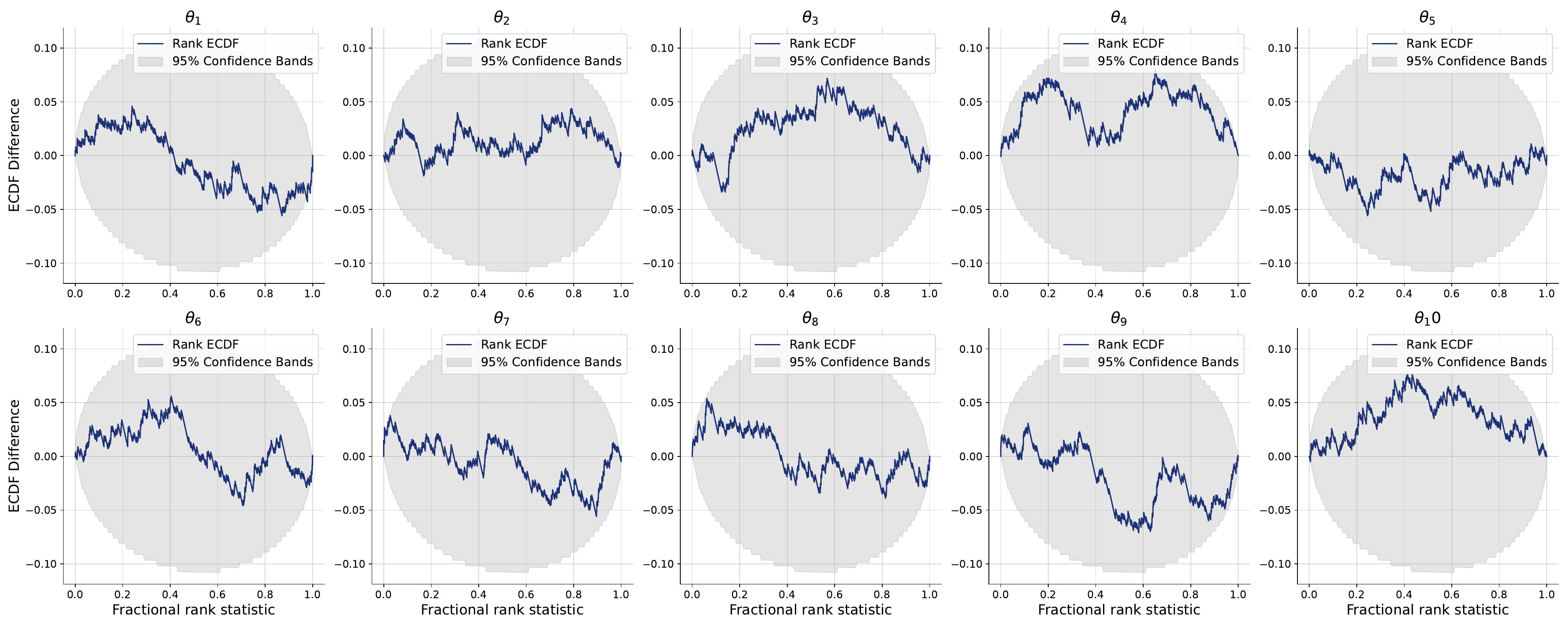}
        \caption{Simulation-based calibration checking.}
    \end{subfigure}
    \caption{Training-phase diagnostics for the Bernoulli GLM problem. The parameter recovery is strong for all parameters. Simulation-based calibration checking indicates good calibration for all parameters. Parameter recovery and simulation-based calibration checking indicate acceptable convergence of the amortized posterior estimator.}
    \label{app:fig:training_diagnostics-BernoulliGLM}
\end{figure}

\paragraph{Training-phase diagnostics.} The closed-world diagnostics (parameter recovery and simulation-based calibration checking) in \autoref{app:fig:training_diagnostics-BernoulliGLM} indicate that the neural network training has successfully converged to an acceptable posterior estimator within the scope of the training set.

\paragraph{Test datasets.}
We generate $K=\num{10000}$ in-distribution test datasets by sampling parameters from the model prior and simulating corresponding observations $\{y_i\}_{i=1}^{100}$ from the Bernoulli distribution.

\subsection{Psychometric curve fitting}

\paragraph{Problem description.}
We adopt an overdispersed psychometric model~\citep{schuttPainfreeAccurateBayesian2016} with the error function (erf) as the sigmoid function in the psychometric function:
\begin{equation}
    \psi(x; m, w, \lambda, \gamma) = \gamma + (1 - \lambda - \gamma)\, \erf(x; m, w),
\end{equation}
where $m$ is the threshold, $w$ is the width, $\lambda$ is the lapse rate, and $\gamma$ is the guess rate.

The full probabilistic model is defined as follows:
\begin{equation}
    \begin{aligned}
        \tilde{m} &\sim \text{Beta}(2, 2), \\
        w &\sim \text{Half-Normal}(0, 1), \\
        \lambda, \gamma, \eta &\sim \text{Beta}(1, 10), \\
        m &= 2\tilde{m} - 1, \\
        \bar{p}_i &= \psi(x_i; m, w, \lambda, \gamma), \\
        p_i &\sim \text{Beta}\left( \left( \frac{1}{\eta^2} - 1 \right) \bar{p}_i,\; \left( \frac{1}{\eta^2} - 1 \right)(1 - \bar{p}_i) \right), \\
        y_i &\sim \text{Binomial}(n_i, p_i),
    \end{aligned}
\end{equation}
where $n_i$ denotes the number of trials, and $x_i$ is the stimulus level. Stimuli are presented at nine fixed levels: $x_i \in \{-100.0,\ -25.0,\ -12.5,\ -6.25,\ 0.0,\ 6.25,\ 12.5,\ 25.0,\ 100.0\}$ and each value is further normalized by dividing by 100.

\paragraph{Simulation budgets.} 
We use \num{50000} simulated parameter–observation pairs for training the amortized estimator, \num{1000} for validation, and \num{200} for training-phase diagnostics, including parameter recovery and simulation-based calibration.

\paragraph{Summary network.}
We use a DeepSet as the summary network, which maps the input dataset to a 16-dimensional summary statistic. The DeepSet has a depth of 2, uses a \emph{gelu} activation, mean inner pooling layers, 64 units in the equivariant and invariant modules, and 5\% dropout.

\begin{figure}[t]
    \centering
    \begin{subfigure}[t]{\linewidth}
        \includegraphics[width=1.0\linewidth]{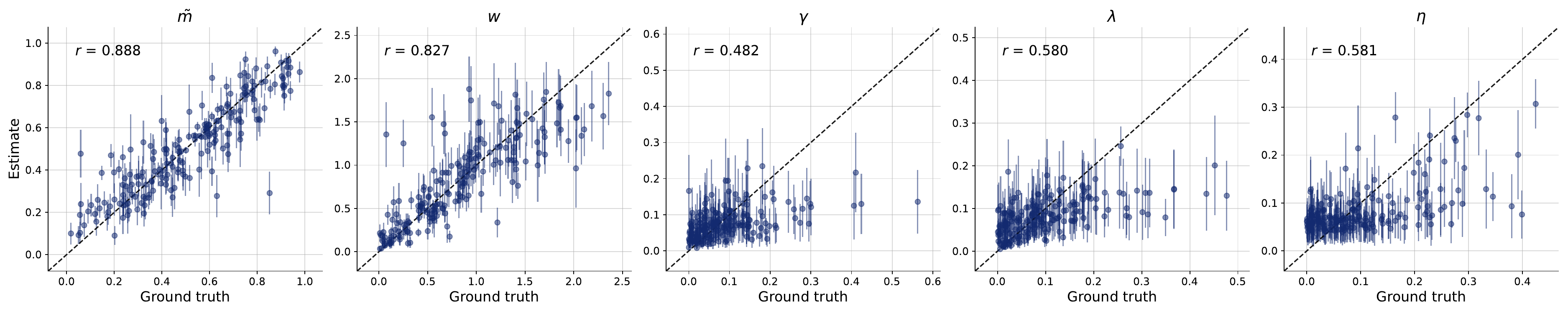}
        \caption{Parameter recovery checking.}
    \end{subfigure}
    \\
    \begin{subfigure}[t]{\linewidth}
        \centering
        \includegraphics[width=1.0\linewidth]{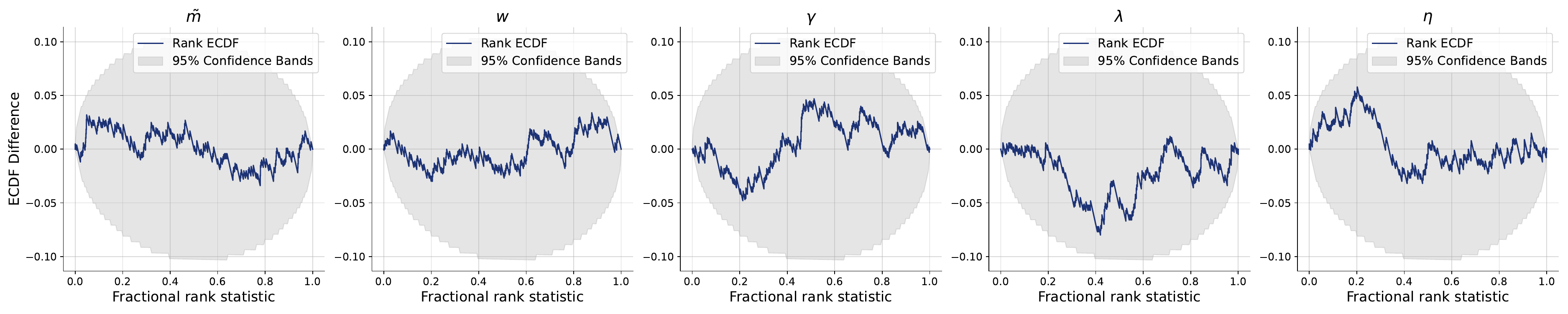}
        \caption{Simulation-based calibration checking.}
    \end{subfigure}
    \caption{Training-phase diagnostics for the psychometric curve fitting problem. Recovery is good for $\tilde{m}$ and $w$, while the other parameters exhibit weaker recoverability. Simulation-based calibration checking indicates excellent calibration for all parameters. Parameter recovery and simulation-based calibration checking indicate acceptable convergence of the amortized posterior estimator.}
    \label{app:fig:training_diagnostics-psychometric}
\end{figure}

\paragraph{Training-phase diagnostics.} The closed-world diagnostics (parameter recovery and simulation-based calibration checking) in \autoref{app:fig:training_diagnostics-psychometric} indicate that the neural network training has successfully converged to an acceptable posterior estimator within the scope of the training set.

\paragraph{Test datasets.}
Our empirical evaluation uses 8,526 mouse behavioral datasets from the International Brain Laboratory public database \citep{theinternationalbrainlaboratoryStandardizedReproducibleMeasurement2021}. We retrieve the data using the provided API with the argument \texttt{task="biasedChoiceWorld"}, which corresponds to behavioral data collected after the mice have completed training. Each dataset is processed into an observation tensor of shape $(9, 3)$, where each row contains the number of correct trials $y_i$, the total number of trials $n_i$, and the stimulus level $x_i$.

\subsection{Decision model}

\paragraph{Problem description.} 
Following \citet{vonkrauseMentalSpeedHigh2022}, we specify the prior distributions for the drift-diffusion model parameters as:
\begin{equation}
    \begin{aligned}
        v_1, v_2 &\sim \text{Gamma}(2, 1), \\
        a_1, a_2 &\sim \text{Gamma}(6, 0.15), \\
        \tau_c &\sim \text{Gamma}(3, 0.15), \\
        \tau_n &\sim \text{Gamma}(3, 0.5),
    \end{aligned}
\end{equation}
where all Gamma distributions use the shape–scale parametrization.\footnote{The prior distributions for the boundary separation parameters $a_1$ and $a_2$ differ slightly from those in \citet{vonkrauseMentalSpeedHigh2022} due to a different parametrization of boundary separation.} We implement the drift-diffusion model likelihood using the \texttt{hssm} package \citep{fengler2025hssm} and PyMC.

\paragraph{Simulation budgets.} 
We use \num{100000} simulated parameter–observation pairs for training the amortized estimator, \num{1000} for validation, and \num{200} for training-phase diagnostics, including parameter recovery and simulation-based calibration.

\paragraph{Summary network.} We use a SetTransformer as the summary network, which maps the input dataset to a 16-dimensional summary statistic. The SetTransformer has two set attention blocks, followed by a pooling multi-head attention block and a fully connected output layer. Each multilayer perceptron (MLP) in the set blocks has two hidden layers of width 128, with \emph{gelu} activation and 5\% dropout.

\begin{figure}[t]
    \centering
    \begin{subfigure}[t]{\linewidth}
        \includegraphics[width=1.0\linewidth]{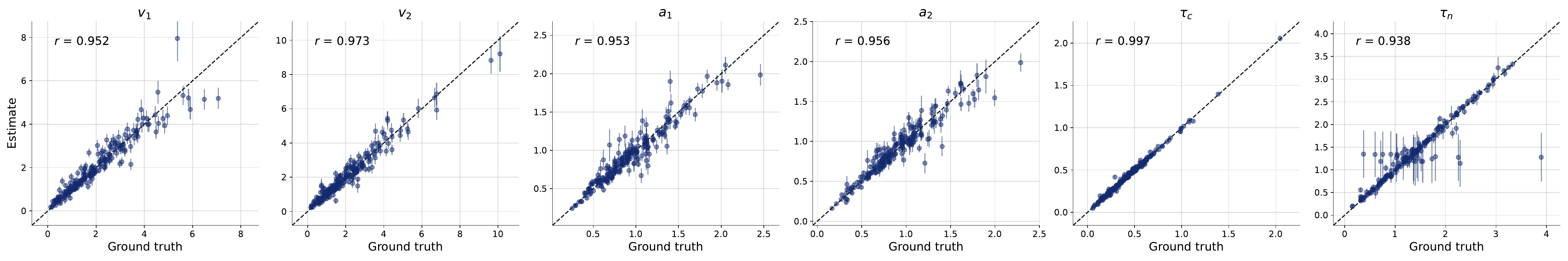}
        \caption{Parameter recovery checking.}
    \end{subfigure}
    \\
    \begin{subfigure}[t]{\linewidth}
        \centering
        \includegraphics[width=1.0\linewidth]{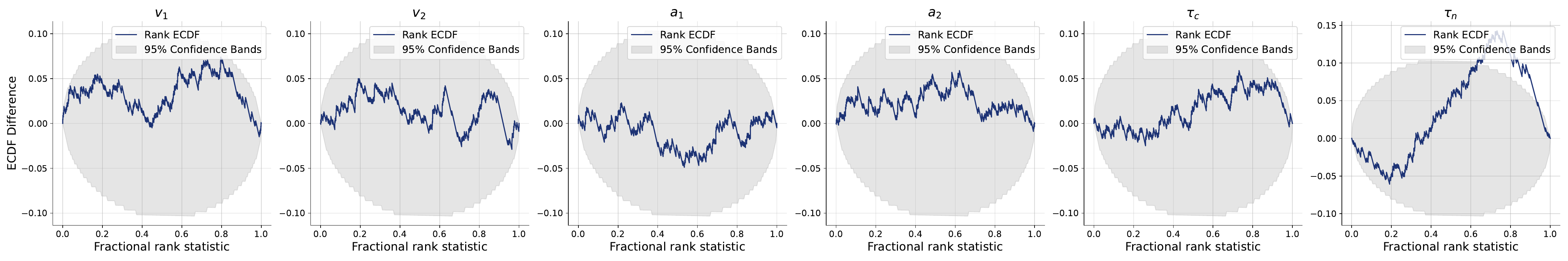}
        \caption{Simulation-based calibration checking.}
    \end{subfigure}
    \caption{Training-phase diagnostics for the decision model. Parameter recovery is strong for all parameters. Simulation-based calibration checking indicates good calibration for all parameters except $\tau_n$, which shows mild deviations, suggesting occasional overestimation by the amortized estimator for this parameter. Parameter recovery and simulation-based calibration checking indicate acceptable convergence of the amortized posterior estimator.}
    \label{app:fig:training_diagnostics-CustomDDM}
\end{figure}

\paragraph{Training-phase diagnostics.} The closed-world diagnostics (parameter recovery and simulation-based calibration checking) in \autoref{app:fig:training_diagnostics-CustomDDM} indicate that the neural network training has successfully converged to an acceptable posterior estimator within the scope of the training set.

\paragraph{Test datasets.}
The test datasets consist of 15,000 participants pre-processed from the online implicit association test (IAT) database \citep{xuPsychologyDataRace2014, vonkrauseMentalSpeedHigh2022}. Each test dataset is a tensor of shape $(120, 4)$, where each row corresponds to a single trial and contains the response time, missing data mask, experiment condition type, and stimulus type.

\section{Additional experimental study of the OOD diagnostic in Step 1}
\label{app:sec:ood_test_study}

\begin{figure}[t]
    \centering
    \begin{subfigure}[t]{\linewidth}
        \centering
        \includegraphics[width=\linewidth]{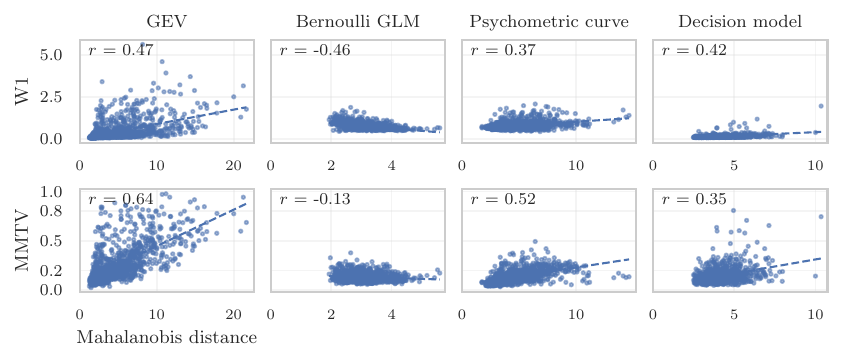}
        \caption{Posterior quality metrics (W1 and MMTV) versus Mahalanobis distance for the four benchmark tasks.}
        \label{app:fig:posterior_metrics_vs_Mahalanobis_dist}
    \end{subfigure}
    \\
    \begin{subfigure}[t]{\linewidth}
        \centering
        \includegraphics[width=\linewidth]{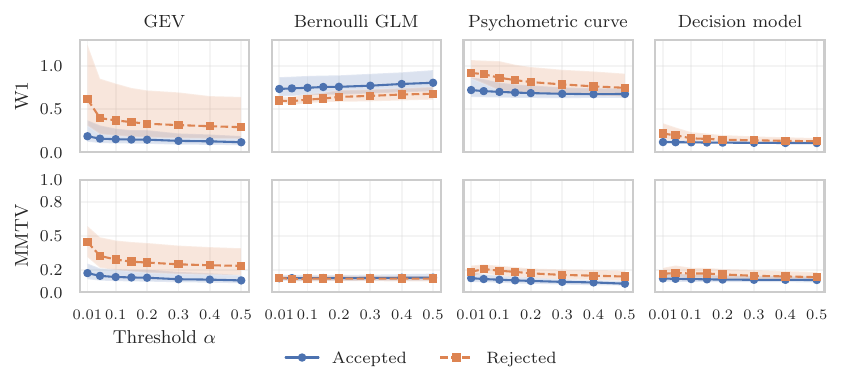}
        \caption{Sensitivity of the rejection threshold $\alpha$ in the OOD test. The median $\pm$ IQR (shaded area) of the posterior quality metrics is shown separately for accepted and rejected datasets.}
        \label{app:fig:ood_threshold_study}
    \end{subfigure}
    \caption{Relationship between amortized posterior quality metrics, Mahalanobis distance, and the OOD rejection threshold in Step~1. (a) Scatter plots of W1 and MMTV versus Mahalanobis distance, with Pearson correlation coefficient $r$ reported in each panel. (b) Effect of varying the threshold $\alpha$ on the posterior quality of accepted and rejected amortized posterior draws. See text in \autoref{app:sec:ood_test_study} for details.}
    \label{app:fig:ood_detailed_study}
\end{figure}

To further investigate the relationship between the Mahalanobis distance in the OOD diagnostic and the quality of the amortized posterior, we visualize this relationship using scatter plots in \autoref{app:fig:posterior_metrics_vs_Mahalanobis_dist} for the four tasks considered in the main text. For each task, we use around 1000 test datasets and compute the Pearson correlation coefficient $r$. The Mahalanobis distance is positively correlated with the two posterior quality metrics (W1 and MMTV) for the GEV, psychometric curve, and decision model tasks, where out-of-distribution test datasets are present. For the Bernoulli GLM, the correlation is negative; here, all test datasets were generated from the same distribution (prior simulations) as the training datasets and the Mahalanobis distance is not informative. From \autoref{app:fig:posterior_metrics_vs_Mahalanobis_dist}, we see a key limitation of the Step-1 OOD diagnostic: the Mahalanobis distance is clearly not a perfect proxy for the posterior quality. In particular, the amortized estimator may still yield low-quality posterior draws on a dataset with a smaller Mahalanobis distance, as also observed in \autoref{fig:metrics_boxplots}. 

We next check the impact of the threshold $\alpha$ in the Step-1 OOD diagnostic by varying it from 0.01 to 0.5, specifically over the set $[0.01, 0.05, 0.1, 0.15, 0.2, 0.3, 0.4, 0.5]$, as shown in \autoref{app:fig:ood_threshold_study}. As $\alpha$ increases, more test datasets are rejected, and the overall posterior quality of accepted amortized posterior draws generally improves as measured by the median and IQR of the posterior metrics (lower posterior metric values indicate higher quality). The quality of rejected amortized posterior draws also improves as $\alpha$ increases, while remaining consistently worse than that of the accepted amortized draws.\footnote{For Bernoulli GLM, where test datasets and training datasets come from the same distribution, we observe a slightly reverse trend as $\alpha$ increases. This is consistent with the corresponding result shown in \autoref{app:fig:posterior_metrics_vs_Mahalanobis_dist} and further suggests that the Mahalanobis distance is not a good measure for posterior quality for in-distribution datasets in this case.} Overall, the posterior quality of the accepted amortized draws, in terms of median and IQR of W1 and MMTV, is not very sensitive to the threshold $\alpha$, and $\alpha=0.05$ appears to be a reasonable default choice. 

These results support the use of the Mahalanobis-distance-based OOD test as a lightweight first-line diagnostic in Step~1: it tends to flag the most problematic datasets and thereby improves the quality of accepted amortized posterior draws at negligible additional cost. At the same time, the residual low-quality posteriors at small Mahalanobis distances underscore that this diagnostic \textit{cannot} guarantee accuracy. For applications that require tighter accuracy guarantees, it is therefore natural to enforce escalation to Step~2 (PSIS) irrespective of the OOD outcome, trading additional computation for a more robust posterior approximation.

\clearpage
\section{Amortized initialization for NUTS}
\label{app:sec:amortized_init_nuts}
\begin{figure}[h]
    \centering
    \includegraphics[width=\linewidth]{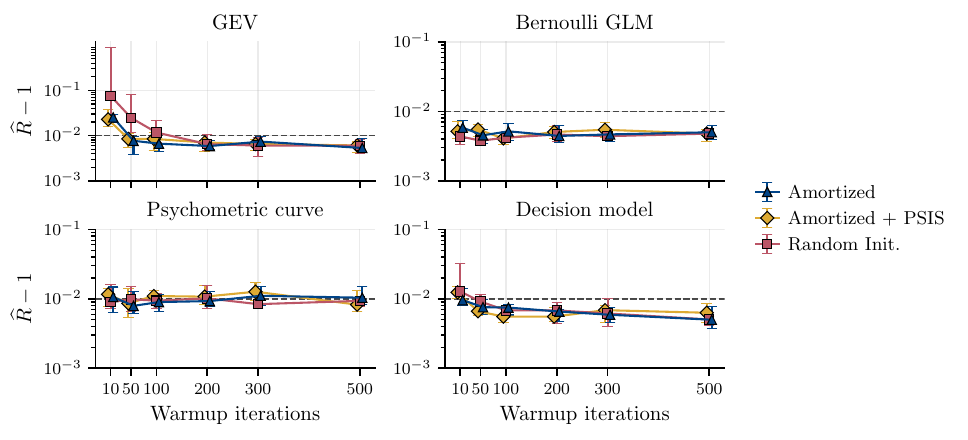}
    \caption{The effect of initialization for NUTS. 
    The figure shows median$\pm$IQR across 20 test datasets. Using amortized posterior draws as initializations for NUTS reduces the required warmup in the GEV and decision model tasks.}
    \label{fig:exp:nuts_inits_comparison}
\end{figure}

In addition to ChEES-HMC, we evaluate the effectiveness of amortized posterior draws 
as initializations for the NUTS sampler. The experimental settings mirror those used for ChEES-HMC (\autoref{sec:amortized_init_chees}), except that we launch only four chains, which is the typical configuration for NUTS.
As shown in \autoref{fig:exp:nuts_inits_comparison}, amortized initializations 
reduce the number of required warm-up iterations for both the GEV problem and the decision model. 
For the psychometric curve and Bernoulli GLM problems, all three initialization methods 
(amortized, PSIS-refined, and random) yield similar convergence behavior according to the $\hat{R}$ diagnostic \citep{Vehtari2021rhat}.

Notably, NUTS generally requires fewer warm-up iterations than ChEES-HMC across the evaluated problems, 
suggesting that while amortized initializations are still beneficial, the relative gain is more pronounced for ChEES-HMC, which runs many short chains in parallel.

\end{document}